\documentclass[manuscript,screen,nonacm]{acmart}
\newcommand{\suppref}[1]{\ref{#1}}

\AtBeginDocument{%
  }

\setcopyright{none}

\acmJournal{TIST}
\acmVolume{0}
\acmNumber{0}
\acmArticle{0}
\acmYear{2026}

\usepackage{amsmath}
\usepackage{graphicx}
\usepackage{booktabs}
\microtypesetup{expansion=false}
\usepackage{comment}
\usepackage{xcolor}
\usepackage{placeins}

\begin{document}

\title{PI-NOMT: Physics-Informed Neural Optimal Mass Transport for Brain Fluid Dynamics}

\author{Mehmet Emin Acar}
\email{acarme22@itu.edu.tr}
\orcid{0009-0006-3046-2072}
\affiliation{%
  \institution{Istanbul Technical University}
  \department{Faculty of Computer and Informatics Engineering, AI and Data Engineering Department}
  \city{Istanbul}
  \country{Turkiye}
}

\author{Vahit Bugra Yesilkaynak}
\email{bugrayesilkaynak@gmail.com}
\orcid{0000-0003-1649-9865}
\affiliation{%
  \institution{Istanbul Technical University}
  \city{Istanbul}
  \country{Turkiye}
}
\author{Helene Benveniste}
\email{helene.benveniste@yale.edu}
\orcid{0000-0002-2887-6667}
\affiliation{%
  \institution{Yale University}
  \department{School of Medicine}
  \city{New Haven, CT}
  \country{USA}
}
\author{Gozde Unal}
\email{gozde.unal@itu.edu.tr}
\orcid{0000-0001-5942-8966}
\affiliation{%
  \institution{Istanbul Technical University}
  \department{Faculty of Computer and Informatics Engineering, AI and Data Engineering Department}
  \city{Istanbul}
  \country{Turkiye}
}

\renewcommand{\shortauthors}{Acar et al.}

\begin{abstract}
Recovering hidden transport mechanisms from sparse spatiotemporal observations is a fundamental inverse problem in scientific machine learning. In brain tracer imaging, dynamic contrast-enhanced MRI (DCE-MRI) provides time-resolved measurements of tracer concentration, while the underlying velocity and source mechanisms governing tracer propagation remain unobserved. We formulate this problem as physics-informed latent-state inference, in which the transport field itself is the primary object of inference rather than an auxiliary variable used only to reconstruct observed densities.

We propose Physics-Informed Neural Optimal Mass Transport (PI-NOMT), a framework that represents density, velocity, and source as continuous neural fields and combines a continuous neural density teacher, recursive differentiable advection--diffusion--source rollout, unbalanced optimal-transport regularization, and governing-equation supervision. The teacher provides continuous interframe targets and derivatives, while the rollout tests whether the inferred latent transport fields can recursively generate the observed tracer evolution. Physical laws act as structural priors that constrain the space of admissible transport mechanisms, while observed tracer dynamics provide evidence for estimating the latent transport state.

We evaluate PI-NOMT on a synthetic benchmark with known ground-truth transport and on DCE-MRI sequences from nine control rats. On the synthetic benchmark, PI-NOMT accurately recovers the prescribed velocity field, including its magnitude, direction, and integrated trajectories, rather than merely reconstructing endpoint densities. Across the nine rat datasets, the framework yields sub-percent local endpoint error, consistent physical speed scales, and low post-training PDE and incompressibility residuals. Ablation studies further show that recursive rollout, physics supervision, and an appropriate magnitude--direction velocity representation play complementary roles in stable latent transport recovery. These results support physics-informed latent-state inference as a general framework for recovering hidden transport mechanisms from observed dynamic scalar fields.

\end{abstract}



\maketitle

\section{Introduction}
\label{sec:intro}

Understanding how substances move through complex biological tissues is
an important problem in neuroscience, physiology, and medical imaging.
In particular, the study of tracer transport through brain tissue has
received considerable attention due to its relevance for fluid
circulation, waste clearance, drug delivery, and the characterization
of neurological disorders. Foundational experimental studies identified a brain-wide paravascular pathway supporting cerebrospinal-fluid exchange with brain tissue and interstitial-solute clearance \cite{iliff2012paravascular}, and subsequently demonstrated its visualization using contrast-enhanced MRI \cite{iliff2013brainwide}.
Dynamic contrast-enhanced magnetic resonance
imaging (DCE-MRI) provides time-resolved observations of tracer
distributions and offers a non-invasive means of studying these
processes. However, while tracer concentrations can be observed over
time, the underlying transport mechanisms responsible for their
evolution remain hidden. Recovering these latent transport processes
from sparse spatiotemporal observations therefore remains a challenging
scientific problem~\cite{bohr2022glymphatic}.

Existing approaches address this problem from several complementary
perspectives. Physics-based methods formulate transport estimation as
an inverse problem governed by fluid transport equations~\cite{toscano26}, while optimal
mass transport (OMT) approaches infer transport fields through
variational optimization over evolving density
distributions~\cite{chen2018regularized,chen2021unbalanced}.
More recently, physics-informed machine learning methods have
demonstrated the effectiveness of incorporating governing equations
into neural network training~\cite{raissi2019physics}. Although these approaches have achieved
promising results, they are typically formulated either as transport
optimization problems, PDE-constrained inverse problems, or predictive
modeling tasks. Here, we take a complementary perspective and formulate the inverse transport problem explicitly as latent-state inference, in which the continuous transport mechanism is represented as the primary latent object to be learned from observations under physical constraints.

We therefore formulate tracer transport recovery as a \emph{latent-state inference problem}. The quantity of scientific
interest is not the observed tracer density itself, but the unobserved
transport field that generates its evolution. Under this viewpoint,
observed tracer distributions provide indirect evidence about an
underlying latent transport state, while governing physical laws
provide structural constraints that restrict the space of admissible
solutions. The objective is therefore not merely to predict future
tracer observations, but to recover a physically meaningful latent
transport field that explains the observed dynamics.

Motivated by this perspective, we propose
\textbf{PI-NOMT} (Physics-Informed Neural Optimal Mass Transport),
a physics-informed latent-state inference
framework for tracer transport modeling. The central idea is to
represent the transport field as an unobserved latent state and to
infer this state directly from tracer observations using transport
physics as a structural prior. To achieve this, PI-NOMT first learns a continuous neural density teacher from the discrete tracer observations, providing interframe supervision and the derivatives required by the governing equations. Neural velocity and source fields are then inferred by requiring them to recursively generate the observed density evolution through a differentiable advection--diffusion--source rollout while satisfying physics and optimal-transport constraints. In this way, rollout serves not merely as a prediction mechanism, but as a test of whether the inferred latent transport state can explain the observed tracer dynamics.

PI-NOMT therefore differs from pairwise OMT-based formulations in how the transport quantities are represented and inferred: rather than re-optimizing interval-specific transport variables, it learns continuous neural transport fields shared across the observed temporal interval and constrains them through recursive physical evolution. Importantly, the goal is not to introduce new biological observables, but to provide a different inference framework for recovering established transport quantities such as velocity fields, transport trajectories, and source distributions. 
In this formulation, physical laws act as structural priors that constrain the latent
representation, while observational data provide evidence about the
underlying transport dynamics. This viewpoint establishes a bridge
between optimal transport, scientific machine learning, and latent
state estimation.

The contributions of this work are summarized as follows:
\begin{itemize}

\item We formulate tracer transport recovery as physics-informed
latent-state inference, in which continuous velocity and source fields
constitute the latent transport state inferred from observed density
evolution under governing physical constraints.

\item We propose PI-NOMT, which combines a continuous neural density
teacher with neural transport fields, unbalanced optimal-transport
regularization, and continuous governing-equation supervision.

\item We develop a recursive differentiable
advection--diffusion--source rollout that tests whether the inferred
latent transport fields can generate the observed tracer evolution,
coupling latent-state estimation directly to physical dynamics.

\item We validate transport-field recovery on a synthetic benchmark
with known ground truth and evaluate cross-subject consistency across DCE-MRI sequences from nine control rats. Ablation studies further characterize the complementary roles of recursive rollout, physics supervision, and velocity-field parameterization in stable transport inference.

\end{itemize}

\section{Related Work}
\label{sec:related}

Optimal Mass Transport (OMT) provides a principled framework for inferring
transport between evolving distributions, from the classical formulations of
Monge~\cite{monge1781memoire} and Kantorovich~\cite{kantorovich1942translocation}
to the dynamic fluid formulation of Benamou and Brenier~\cite{benamou2000computational}.
Extensions to unbalanced transport jointly represent spatial transport and mass
variation, including Wasserstein--Fisher--Rao and related dynamic formulations
\cite{chizat2018interpolating,chizat2018unbalanced}, while numerical schemes have
addressed general advection--reaction--diffusion settings
\cite{lombardi2015eulerian,gallouet2019splitting}.
For biological tracer transport, Regularized OMT (rOMT) incorporates diffusion
into the transport dynamics~\cite{chen2018regularized}, and Unbalanced
Regularized OMT (urOMT) additionally introduces spatially varying source and
sink terms~\cite{chen2021unbalanced}. These formulations have enabled estimation
and visualization of glymphatic transport, tracer propagation, and dynamic
influx and clearance in neuroimaging studies
\cite{koundal2020optimal,koundal2024divergent}. PI-NOMT builds on this
velocity--source--diffusion perspective but differs in the inference paradigm:
rather than optimizing interval-specific transport variables over observed
density trajectories, it learns shared continuous neural velocity and source
fields whose validity is tested through recursive physical evolution.

Scientific Machine Learning (SciML) provides a complementary route for
recovering hidden physical quantities from indirect observations. Physics-Informed
Neural Networks (PINNs) incorporate governing-equation residuals into neural
optimization~\cite{raissi2019physics}, while Hidden Fluid Mechanics demonstrated
recovery of latent velocity and pressure fields from time-resolved scalar
observations~\cite{raissi2020hidden}. Related physics-informed AI velocimetry
has reconstructed three-dimensional cerebrospinal-fluid velocity fields from
sparse measurements~\cite{toscano2024inferring}. More broadly, differentiable
simulation enables gradients to propagate through physical solvers and thereby
couples state estimation directly to simulated dynamics
\cite{degrave2019differentiable}. PI-NOMT follows this physics-informed inverse
modeling perspective, but combines continuous latent transport fields,
unbalanced optimal-transport regularization, and recursive differentiable
advection--diffusion--source rollout within a single inference framework.

Brain transport estimation from dynamic contrast-enhanced MRI (DCE-MRI) has
been approached using both mechanistic inverse models and optimal-transport
methods. MR-AIV estimates velocity, pressure, and permeability fields from
DCE-MRI under Darcy-flow assumptions~\cite{toscano26}, while PINN-based methods
have estimated parameters of mechanistic molecular-transport models from
human-brain MRI~\cite{zapf2022investigating}. Optimal-transport approaches range
from early OMT-based estimation of glymphatic transport
\cite{ratner2015optimal,ratner2017cerebrospinal} to regularized visualization
and inference of glymphatic pathways
\cite{elkin2018optimal,koundal2020optimal}. These studies establish that
physiologically meaningful transport quantities can be inferred from dynamic
imaging under appropriate physical assumptions. PI-NOMT addresses the related
but distinct problem of representing spatially and temporally varying velocity
and source fields as a shared latent transport state whose consistency with the
observations is tested through recursive advection--diffusion--source dynamics.

A related line of work learns latent continuous-time dynamics directly from
partial observations. Neural ODEs parameterize continuous dynamics with neural
networks~\cite{chen2018neuralode}, while latent state-space and sequential models
infer hidden states from incomplete observations
\cite{rangapuram2018deepstatespace,karl2017deepvariational}. These approaches
establish latent-state inference as a general computational paradigm, but do not
specifically address recovery of continuous velocity--source fields from
tracer-density observations under unbalanced transport physics. PI-NOMT brings
these perspectives together by treating the transport mechanism itself as the
latent state and constraining its inference through observed density evolution,
recursive differentiable rollout, optimal-transport regularization, and
governing-equation supervision. To the best of our knowledge, prior work has
not combined these elements to formulate DCE-MRI transport-field recovery as
continuous physics-informed latent-state inference.

\section{Methodology}
\label{sec:methodology}

PI-NOMT formulates brain tracer analysis as physics-informed latent-state
inference. The observations are a sequence of 3D tracer-density volumes, while
the latent state of interest is the transport mechanism that generates their
evolution. PI-NOMT first fits a continuous neural density field to the measured
sequence and freezes it as a differentiable teacher. Continuous velocity and
source fields are then inferred by requiring them to recursively generate the
teacher density through differentiable advection--diffusion--source dynamics
while satisfying optimal-transport and governing-equation constraints.

\subsection{Latent Transport Formulation}

Let $\Omega\subset\mathbb{R}^3$ denote the analysis domain and
$\rho^{obs}(\mathbf{x},t_i)$ the measured tracer density at discrete frame
times $t_i$. We define the latent transport state as
$\{\mathbf{v}(\mathbf{x},t),r(\mathbf{x},t)\}$, where
$\mathbf{v}$ is the velocity field and $r$ is a relative source or sink rate.

PI-NOMT builds on dynamic unbalanced optimal transport, in which spatial
transport and local mass variation are coupled through the objective
\begin{equation}
    \min_{\rho,\mathbf{v},r}
    \int_{t_a}^{t_b}\int_{\Omega}
    \rho\|\mathbf{v}\|_2^2+\alpha\rho r^2
    \,d\mathbf{x}\,dt
    +
    \beta\int_\Omega
    \left(
    \rho(\mathbf{x},t_b)-\rho^{obs}(\mathbf{x},t_b)
    \right)^2d\mathbf{x},
    \label{eq:uromt_objective}
\end{equation}
subject to the advection--diffusion--source equation
\begin{equation}
    \frac{\partial\rho}{\partial t}
    +\nabla\cdot(\rho\mathbf{v})
    =
    D\nabla^2\rho+\rho r ,
    \label{eq:adr_conservative}
\end{equation}
where $D$ is a fixed scalar diffusion coefficient
\cite{chen2021unbalanced}.

PI-NOMT retains this velocity--source--diffusion decomposition but changes the
representation and inference procedure. Rather than estimating independent
voxelwise transport variables for successive observed intervals, it represents
the velocity and source as shared continuous neural fields
$\mathbf{v}_{\boldsymbol{\theta}}(\mathbf{x},t)$ and
$r_{\boldsymbol{\theta}}(\mathbf{x},t)$ over the full temporal domain. Their
inference is coupled to the observed density evolution through the recursive
differentiable physical rollout described below and constrained continuously
through the governing equation.

\subsection{Continuous Neural Density and Transport Fields}

The first stage of PI-NOMT fits a continuous neural representation of the
observed tracer density,
\begin{equation}
    \rho_\phi(\mathbf{x},t)
    =
    \mathcal{N}_{\rho}(\mathbf{x},t;\boldsymbol{\phi}),
    \label{eq:rho_network}
\end{equation}
from tracer measurements at the discrete acquisition times. After fitting,
$\rho_\phi$ is frozen throughout transport inference. It therefore provides a
fixed continuous reference that can be queried at both measured and interframe
times, as well as the automatic-differentiation quantities
$\partial_t\rho_\phi$, $\nabla\rho_\phi$, and $\nabla^2\rho_\phi$ required by
the physics objective. The teacher is learned from the observed tracer data
independently of the transport model and is not itself constrained by the
transport PDE. Network architecture and density-teacher training details are
given in Supplementary Sec.~\suppref{supp:network_details}. 

The second stage represents the relative source and velocity as continuous
neural fields. The source is
\begin{equation}
    r_{\boldsymbol{\theta}}(\mathbf{x},t)
    =
    \mathcal{N}_{s}(\mathbf{x},t;\boldsymbol{\theta}_s),
    \label{eq:source_def}
\end{equation}
while the velocity is factorized into magnitude and direction,
\begin{equation}
\begin{aligned}
    \tilde{m}_{\boldsymbol{\theta}}
        &=\mathcal{N}_{m}(\mathbf{x},t;\boldsymbol{\theta}_m),\\
    \tilde{\mathbf{d}}_{\boldsymbol{\theta}}
        &=\mathcal{N}_{d}(\mathbf{x},t;\boldsymbol{\theta}_d),\\
    m_{\boldsymbol{\theta}}
        &=\operatorname{softplus}
          (\tilde{m}_{\boldsymbol{\theta}}),\\
    \mathbf{d}_{\boldsymbol{\theta}}
        &=\frac{\tilde{\mathbf{d}}_{\boldsymbol{\theta}}}
        {\|\tilde{\mathbf{d}}_{\boldsymbol{\theta}}\|_2+\epsilon},\\
    \mathbf{v}_{\boldsymbol{\theta}}
        &=m_{\boldsymbol{\theta}}\mathbf{d}_{\boldsymbol{\theta}} .
\end{aligned}
\label{eq:split_velocity}
\end{equation}
The nonnegative magnitude and normalized direction branches separate transport
speed from directional structure. The magnitude branch receives a Chebyshev
feature expansion to represent localized variations in transport speed, whereas
the direction branch uses the spatiotemporal coordinates directly, encoding the
inductive bias that transport direction varies more smoothly across neighboring
regions. The effect of this factorization is evaluated against a Cartesian
velocity parameterization in Table~\ref{tab:transport_tradeoffs}. PI-NOMT-Cart retains the same learning framework but predicts the three Cartesian velocity components directly. Detailed network architectures are provided in Supplementary Sec.~\suppref{supp:network_details}. 

\subsection{Differentiable Advection--Diffusion--Source Rollout}
\label{sec:rollout}

The inferred transport fields are evaluated through a differentiable recursive
rollout that tests whether they can generate the observed density evolution
under Eq.~\eqref{eq:adr_conservative}. For an interval $[t_a,t_b]$ divided into
$K$ substeps of size $\Delta t=(t_b-t_a)/K$, the current density
$\hat{\rho}^{k}$ is updated sequentially by source, advection, and diffusion.

The relative-source update is
\begin{equation}
    \hat{\rho}^{k,\mathrm{src}}(\mathbf{x})
    =
    \left(1+\Delta t\,r^k(\mathbf{x})\right)
    \hat{\rho}^{k}(\mathbf{x}),
    \label{eq:source_step}
\end{equation}
where $r^k(\mathbf{x})$ denotes the source field at the current substep.

Advection is implemented with a differentiable semi-Lagrangian backtrace,
\begin{equation}
\begin{aligned}
    \mathbf{x}_d
    &=
    \mathbf{x}-\Delta t\,\mathbf{v}^k(\mathbf{x}),\\
    \hat{\rho}^{k,\mathrm{adv}}(\mathbf{x})
    &=
    \mathcal{I}
    \!\left[\hat{\rho}^{k,\mathrm{src}}\right](\mathbf{x}_d),
\end{aligned}
\label{eq:advection_step}
\end{equation}
where $\mathcal{I}$ denotes differentiable trilinear sampling.

The characteristic update above corresponds to
$\rho_t+\mathbf{v}\!\cdot\!\nabla\rho=0$. Since
$\nabla\!\cdot(\rho\mathbf{v})
=\mathbf{v}\!\cdot\!\nabla\rho+\rho\nabla\!\cdot\mathbf{v}$,
it is consistent with the conservative advection equation under the
approximately incompressible assumption
$\nabla\!\cdot\mathbf{v}\approx0$ adopted here. 
The characteristic update is therefore consistent with the conservative form under the approximate-incompressibility assumption adopted here; deviations from this assumption are handled by the divergence penalty introduced in Section~\ref{sec:physics_losses}.

Diffusion is then applied by a backward-Euler update,
\begin{equation}
    \left(I-\Delta t D\nabla_h^2\right)
    \hat{\rho}^{k,\mathrm{diff}}
    =
    \hat{\rho}^{k,\mathrm{adv}},
    \label{eq:diffusion_step}
\end{equation}
after which
$\hat{\rho}^{k+1}=\hat{\rho}^{k,\mathrm{diff}}$ becomes the input to the next
source--advection--diffusion cycle. The rollout is therefore fully recursive:
the density is initialized only at the beginning of a rollout window and is not
reset to the teacher between substeps. Unless otherwise stated, PI-NOMT uses
$K=10$ substeps per sampled rollout interval.

Implementation details of the implicit diffusion solve, boundary treatment, and
its differentiable backward pass are provided in Supplementary
Sec.~\suppref{supp:rollout_implementation}. 

\subsection{Physics Supervision}
\label{sec:physics_losses}

Physics supervision constrains the inferred transport fields using the frozen
continuous density teacher. Under the approximately incompressible formulation
used by the rollout, Eq.~\eqref{eq:adr_conservative} yields the pointwise
residual
\begin{equation}
R_{\mathrm{PDE}}
=
\partial_t \rho_\phi
+
\mathbf{v}_{\boldsymbol{\theta}}\cdot\nabla\rho_\phi
-
D\nabla^2\rho_\phi
-
\rho_\phi r_{\boldsymbol{\theta}} .
\label{eq:physics_residual}
\end{equation}
Deviations from incompressibility are penalized separately through
$\nabla\!\cdot\mathbf{v}_{\boldsymbol{\theta}}$.

The quantities
$\partial_t\rho_\phi$, $\nabla\rho_\phi$, and $\nabla^2\rho_\phi$
are obtained by automatic differentiation of the frozen density teacher at
sampled space--time locations. Consequently, the physics objective evaluates a
continuous residual of the observed density representation and the inferred
latent fields rather than a finite-difference residual of the rolled-out
density. Gradients from this supervision update the velocity and source
networks but do not modify the frozen teacher.

\subsection{Training Objective}
\label{sec:training_procedure}

Transport-field inference uses random temporal windows. Each window is
initialized from the frozen density teacher and propagated recursively using
the differentiable source--advection--diffusion update of
Section~\ref{sec:rollout}. The nominal rollout contains $K=10$ substeps, and
the teacher provides continuous target densities
$\rho_\phi^k=\rho_\phi(\cdot,t_k)$ at all sampled times. Thus, reconstruction
supervision is available both at the endpoint and at interframe rollout states.
Temporal-window sampling and boundary handling are detailed in Supplementary Sec.~\suppref{supp:training_details}. 

\begin{figure*}[t]
    \centering
    \includegraphics[width=0.96\textwidth]{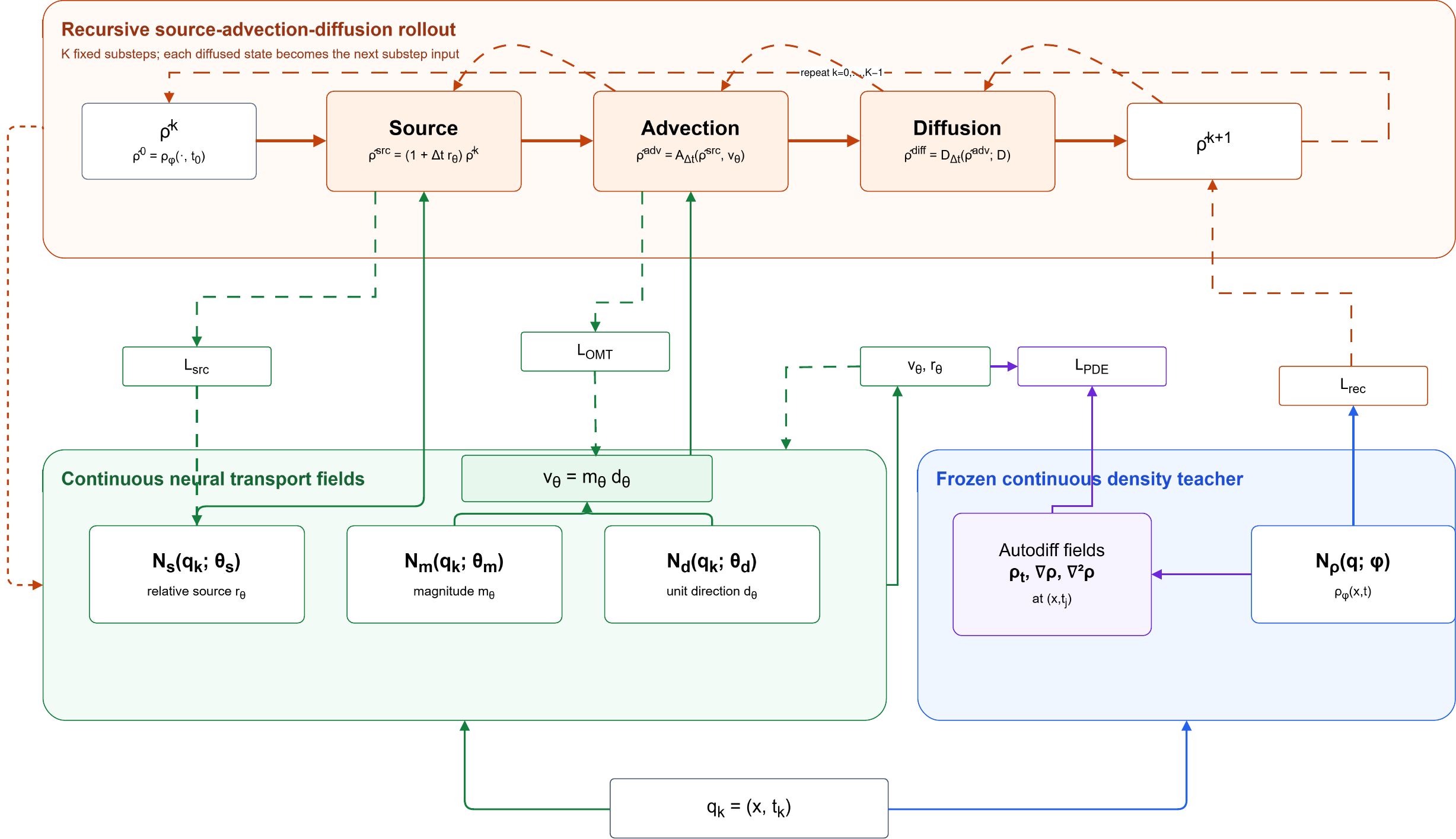}
    \caption{PI-NOMT training pipeline. The frozen density teacher provides
    rollout initialization, continuous reconstruction targets, and derivatives
    for physics supervision. The learned source and velocity fields generate a
    recursive differentiable rollout, with gradients supplied by reconstruction,
    optimal-transport, source, PDE, and incompressibility objectives.}
    \Description{PI-NOMT training diagram showing a frozen continuous density
    teacher, trainable source and magnitude--direction velocity fields, and a
    recursive source--advection--diffusion rollout.}
    \label{fig:pi_nomt_pipeline}
\end{figure*}

Figure~\ref{fig:pi_nomt_pipeline} summarizes the resulting training pipeline.
The frozen teacher provides rollout initialization, reconstruction targets, and
the derivatives used by the physics residual, while the trainable velocity and
source fields are optimized through reconstruction, optimal-transport,
source, and physics objectives.

The endpoint and intermediate reconstruction losses are
\begin{align}
    \mathcal{L}_{final}
    &=
    \operatorname{MSE}_{\Omega}
    \left(\hat{\rho}^{K},\rho_\phi^K\right),
    \\
    \mathcal{L}_{inter}
    &=
    \frac{1}{K-1}
    \sum_{k=1}^{K-1}
    \operatorname{MSE}_{\Omega}
    \left(\hat{\rho}^{k},\rho_\phi^k\right).
    \label{eq:reconstruction_losses}
\end{align}
Both terms are computed from the same recursive rollout, so their gradients
propagate through the sequence of learned source and velocity fields rather
than through independently teacher-initialized transitions.

Optimal-transport and source regularization are evaluated along the predicted density trajectory,
\begin{align}
    \mathcal{L}_{omt}
    &=
    \frac{1}{K}\sum_{k=0}^{K-1}
    \mathbb{E}_{\mathbf{x}\in\Omega}
    \left[
    \operatorname{sg}(\hat{\rho}^{k+1})
    \|\mathbf{v}^{k}\|_2^2
    \right],
    \label{eq:omt_loss}
    \\
    \mathcal{L}_{src}
    &=
    \frac{1}{K}\sum_{k=0}^{K-1}
    \mathbb{E}_{\mathbf{x}\in\Omega}
    \left[
    \operatorname{sg}(\hat{\rho}^{k+1})
    (r^{k})^2
    \right].
    \label{eq:source_loss}
\end{align}
where $\operatorname{sg}(\cdot)$ denotes stop-gradient, so the rolled density
acts as a weighting measure without receiving gradients from these
regularizers.

Physics supervision is applied at the sampled substep times,
\begin{align}
    \mathcal{L}_{phys}
    &=
    \frac{1}{K}\sum_{k=0}^{K-1}
    \mathbb{E}_{\mathbf{x}\in\Omega}
    \left[
    R_{PDE}(\mathbf{x},t_k)^2
    \right],
    \\
    \mathcal{L}_{inc}
    &=
    \frac{1}{K}\sum_{k=0}^{K-1}
    \mathbb{E}_{\mathbf{x}\in\Omega}
    \left[
    \left(
    \nabla\cdot
    \mathbf{v}_{\boldsymbol{\theta}}(\mathbf{x},t_k)
    \right)^2
    \right].
    \label{eq:physics_losses}
\end{align}

The complete PI-NOMT objective is
\begin{equation}
    \mathcal{L}
    =
    w_{final}\mathcal{L}_{final}
    + w_{inter}\mathcal{L}_{inter}
    + w_{omt}\mathcal{L}_{omt}
    + w_{src}\mathcal{L}_{src}
    + w_{phys}\mathcal{L}_{phys}
    + w_{inc}\mathcal{L}_{inc}.
    \label{eq:total_loss}
\end{equation}
The rollout state is not detached between substeps; reconstruction gradients
therefore propagate through the complete recursive transport chain within each
sampled window. Additional optimization and gradient-computation details are
provided in Supplementary Sec.~\suppref{supp:training_details}. 
\section{Experimental Evaluation}

We evaluate PI-NOMT on a synthetic Gaussian-sphere benchmark and a cohort of 3D rat-brain tracer sequences.  
The synthetic benchmark provides known ground-truth transport dynamics and therefore enables direct evaluation of latent-field recovery, whereas the rat-brain experiments assess reconstruction, recursive transport behavior, and the physical plausibility of inferred transport fields when ground-truth velocity and source fields are unavailable.
In both settings, evaluation follows the two-stage PI-NOMT procedure: the continuous density field is first fitted and frozen, after which the latent transport fields are learned through recursive physical rollout.

\subsection{Dataset, Models, and Metrics}

\subsubsection{Synthetic Gaussian-Sphere Benchmark}
\label{subsec:SynthGaussian}
The synthetic experiment uses five $50\times50\times50$ Gaussian-sphere
volumes at times $t=0,\ldots,4$.  In the analytic coordinates used to generate the volumes, the Gaussian center is translated by $(0.8,0.8,0.8)$ per frame. Its central amplitude is scaled by $[1.0,1.1,1.2,1.1,1.0]$, producing mass gain over the first two intervals and mass loss over the final two, and increasing Gaussian smoothing is applied over time.  In addition to the five training volumes, we generate 36 held-out volumes at increments of $0.1$ using the same continuous center, amplitude, and smoothing rules.  These 36 interframe volumes are withheld from density-teacher fitting and are used exclusively to evaluate its temporal interpolation accuracy.
Under the analytic-coordinate-to-voxel-grid mapping and the stored voxel-axis orientation, the prescribed analytic translation corresponds to a displacement of $(-3.27,-3.27,-3.27)$ voxels per frame interval. Since each frame interval spans four numerical time units, the resulting constant reference velocity is 
$\mathbf{v}_{\mathrm{GT}}=(-0.8167,-0.8167,-0.8167)$ voxels per numerical time unit.

Given the spatially constant reference velocity, we use the Cartesian PI-NOMT variant (PI-NOMT-Cart) for this benchmark. The magnitude--direction representation is evaluated in the more challenging rat-brain experiments, where spatially heterogeneous transport provides a meaningful setting for evaluating its structured inductive bias. The density, source, and velocity networks each use two residual blocks of width 32 with $\tanh$ activations.  The source branch and the
same differentiable source-advection-diffusion rollout are retained.

\noindent For a prediction $\hat{\rho}(\cdot,t)$ and reference
$\rho_{ref}(\cdot,t)$, normalized mean squared error (NMSE, \%) is reported as
\begin{equation}
    \operatorname{NMSE}(t)
    =
    100
    \frac{
    \left\|\hat{\rho}(\cdot,t)-\rho_{ref}(\cdot,t)\right\|_2^2
    }{
    \left\|\rho_{ref}(\cdot,t)\right\|_2^2
    } .
    \label{eq:nmse_metric}
\end{equation}

For the Gaussian benchmark, we define the evaluation support as
$\mathcal{S}(t)=\{\mathbf{x}:\|\mathbf{x}-\mathbf{c}_{\mathrm{GT}}(t)\|_2
\leq 2\sigma_0\}$, a fixed-radius sphere centered at the prescribed Gaussian
center $\mathbf{c}_{\mathrm{GT}}(t)$, where $\sigma_0$ denotes the initial
Gaussian standard deviation. We report 
\begin{align}
    \mathcal{E}_{\mathrm{EPE}}
    &=\frac{1}{N_t}\sum_t\frac{1}{|\mathcal{S}(t)|}
      \sum_{\mathbf{x}_i\in\mathcal{S}(t)}
      \left\|\mathbf{v}_{\boldsymbol{\theta}}(\mathbf{x}_i,t)
      -\mathbf{v}_{\mathrm{GT}}\right\|_2, \\
    \mathcal{E}_{\mathrm{mag}}
    &=\frac{1}{N_t}\sum_t\frac{1}{|\mathcal{S}(t)|}
      \sum_{\mathbf{x}_i\in\mathcal{S}(t)}
      \left|\left\|\mathbf{v}_{\boldsymbol{\theta}}(\mathbf{x}_i,t)\right\|_2
      -\left\|\mathbf{v}_{\mathrm{GT}}\right\|_2\right|,
    \label{eq:synthetic_velocity_metrics}
\end{align}
where $N_t$ denotes the number of evaluation times.
Cosine similarity and angular error (degrees) are computed from the same pointwise vector comparisons. Pathline seeds are sampled in proportion to the
initial Gaussian density and retained within $\mathcal{S}(0)$. Analytic and learned advective pathlines are integrated from identical seeds across the four frame intervals with 40 substeps. Pathline EPE is the mean Euclidean distance between corresponding final endpoints, reported in voxels. For density interpolation, we track the sphere center, computed as the density-weighted mean voxel location.  Centroid error is the Euclidean distance between the teacher-predicted center and the known analytic center, in voxels.

\subsubsection{Rat-Brain DCE-MRI Experiment}

The rat-brain experiments use 3D DCE-MRI sequences acquired from nine
three-month-old control rats after gadoteric acid was injected into the
cerebrospinal fluid under anesthesia~\cite{chen2022cerebral}.  The acquisition
protocol samples one volume every 5 minutes.  For consistent analysis across subjects, we retain the 29 preprocessed acquisition frames indexed 6--34 and restrict the analysis to a cropped anatomical brain mask. The mask is morphologically closed, hole-filled, and the resulting volume is cropped to its bounding box.
The cohort comprises C1189, C1191, C1192, C1204, C1213, C1217, C1263, C1264, and C1265.  C1217 is used for the detailed analyses and ablations, while the remaining sequences provide a cross-subject evaluation.  Each MRI series was converted to percentage signal change relative to the pre-contrast baseline, which was used as a surrogate for tracer concentration.  

Following the preprocessing protocol used in urOMT~\cite{chen2021unbalanced},
negative concentrations within this region are set to zero, values above
10,000 are clipped, and ten iterations of affine-invariant nonlinear
curvature-flow smoothing are applied with a step size of 0.1. The processed
concentrations are then scaled to $[0,100]$ using one global maximum computed
across all acquisition frames, rather than independently normalizing each
frame.  Thus, the measured reference used throughout the
experiments is the consistently preprocessed tracer sequence, not the raw MRI
intensity volume.  Each retained voxel is represented by its spatial
coordinates, acquisition time, and normalized tracer concentration.  The
density-teacher experiment covers the complete 29-frame interval, with every third volume reserved as a full-frame validation target and excluded from teacher fitting.  
The nominal transport rollout spans two acquisition-frame intervals, corresponding to a 10-minute endpoint horizon, and is discretized into 10 numerical rollout substeps.
Unless otherwise stated, all quantitative metrics are computed over voxels within this same anatomical mask.

Each 5-minute acquisition interval is represented by two numerical time
units. With the reported $0.300$-mm voxel size~\cite{chen2022cerebral}, a
velocity of one voxel per numerical time unit corresponds to $0.12$~mm/min.
This conversion is used for all reported rat-brain speed maps, trajectory
colors, distributions, and tabulated speed statistics. Flux-vector magnitude
denotes endpoint displacement rather than speed and is converted from voxels to millimeters by multiplying by $0.300$.

Each rat sequence is processed using the complete two-stage PI-NOMT procedure of Section~\ref{sec:methodology}. Unless otherwise stated, PI-NOMT uses the source branch, magnitude--direction velocity parameterization, 10 rollout substeps per two-frame interval, random-window training, and PDE and incompressibility supervision.

Each rat-brain experiment was run on a single NVIDIA RTX 6000 Ada GPU. For the representative full PI-NOMT experiment, fitting the continuous density teacher required 59 min and subsequent training of the transport model required 37 h 23 min, for a total two-stage fitting time of approximately 38 h 22 min. Once trained, post-training evaluation and output generation—including recursive rollout evaluation, Eulerian velocity and source fields, physics diagnostics, pathline/speed-line/Péclet-line computation, and basic rendering—required approximately 10 min 23 s.

We use different labels for the transport variants throughout the experiments.
PI-NOMT denotes the full model with the source branch, magnitude--direction velocity parameterization, and PDE and incompressibility supervision.
PI-NOMT-NoPhys removes the PDE and incompressibility losses while retaining that architecture.  PI-NOMT-NoSrc removes only the source branch. PI-NOMT-Cart replaces the split velocity representation with one width-182 network that directly predicts the three velocity components.  This width approximately matches the hidden-layer parameter budget of the two width-128 magnitude and direction branches. PI-NOMT-2S uses two rollout substeps instead of 10. PI-NOMT-OneStep replaces the recursively supervised rollout objective with local one-step training.  To separate each structural choice from the effect of physics supervision, we evaluate PI-NOMT, NoSrc, Cart, 2S, and OneStep both with and without the PDE and incompressibility losses.

We use two rollout error protocols.  Continuous rollout NMSE initializes the
transport model once at frame 6 and recursively evolves the density to later
frames without resetting.  Reference-start endpoint NMSE evaluates shorter
two-frame chunks, where each chunk starts from the density teacher at the chunk start and is compared with the teacher at the chunk endpoint.  The first metric tests long-horizon accumulation; the second tests local transport accuracy.

For an interval $[t_a,t_b]$, the Eulerian speed map reports the pointwise
temporal average of the instantaneous velocity magnitude,
\begin{equation}
    \overline{s}_{E}(\mathbf{x};t_a,t_b)
    =
    \frac{1}{K}\sum_{k=0}^{K-1}
    \left\|\mathbf{v}_{\boldsymbol{\theta}}(\mathbf{x},t_k)\right\|_2 .
    \label{eq:eulerian_speed_metric}
\end{equation}
It is therefore neither a midpoint sample nor a path-integrated distance.
As a complementary Lagrangian summary, we define the pathline-sampled mean
advective speed as
\begin{equation}
    \bar{s}_{L}
    =
    \frac{1}{N_L}
    \sum_{j=1}^{N_L}
    \left\|
    \mathbf{v}_{\boldsymbol{\theta}}(\mathbf{x}_j,t_j)
    \right\|_2 ,
    \label{eq:pathline_speed_metric}
\end{equation}
where $(\mathbf{x}_j,t_j)$ denotes a mask-inside trajectory sample and $N_L$
is the total number of such samples. Pathlines are initialized from seed points
inside the anatomical mask and integrated using the diffusion-augmented drift
\[
    \mathbf{v}_{\boldsymbol{\theta}}
    -
    D\nabla\log\hat{\rho}.
\]
The reported pathline-sampled speed is the mean local advective magnitude  $\|\mathbf{v}_{\boldsymbol{\theta}}\|_2$ over all mask-inside trajectory-time samples, providing a pooled summary of the advective-speed scale encountered along the integrated trajectories. Each in-domain sample therefore contributes equally; trajectories are not first normalized to have equal weight.

The same integrated trajectories are used for the Lagrangian visualizations. Speed-lines color each trajectory sample by its local instantaneous advective speed $\|\mathbf{v}_{\boldsymbol{\theta}}\|_2$
\cite{koundal2020optimal}, rather than by trajectory-averaged speed or
cumulative path length. Flux vectors summarize the endpoint displacement and direction of the trajectories; their magnitude is reported in millimeters rather than as a speed.

For P\'eclet-line visualization, the local balance between advective and
diffusive transport is defined as~\cite{chen2021unbalanced}
\begin{equation}
    Pe(\mathbf{x},t)
    =
    \frac{
    \|\mathbf{v}_{\boldsymbol{\theta}}(\mathbf{x},t)\|_2
    }{
    D\|\nabla\log\hat{\rho}(\mathbf{x},t)\|_2+\varepsilon
    },
    \label{eq:peclet_metric}
\end{equation}
where $\hat{\rho}$ is the current rolled-out density, and $\varepsilon$ is a numerical stabilizer. Ignoring this stabilizer, the expression corresponds to the local form
$Pe=\|\mathbf{v}\|_2L/D$ with
$L=1/\|\nabla\log\hat{\rho}\|_2$.
P\'eclet-lines use the same diffusion-augmented trajectories and color each trajectory sample by the local $Pe$. Thus, trajectory geometry is determined by the diffusion-augmented drift, whereas speed and P\'eclet coloring retain the corresponding local advective quantities.

Latent-field diagnostics are computed after training using exactly the same
space--time samples for every model.  We evaluate all $N_x=75{,}360$ spatial
points inside the brain mask at $N_t=140$ uniformly spaced times,
$t_j=6.2,6.4,\ldots,34.0$.  At each of these $10{,}550{,}400$ space--time
samples, we compute the PDE residual and velocity divergence.  We then report
their mean squared values:
\begin{equation}
    \mathcal{E}_{\mathrm{PDE}}
    =
    \frac{1}{N_tN_x}
    \sum_{j=1}^{N_t}\sum_{i=1}^{N_x}
    \left|R_{PDE}(\mathbf{x}_i,t_j)\right|^2,
    \qquad
    \mathcal{E}_{\mathrm{inc}}
    =
    \frac{1}{N_tN_x}
    \sum_{j=1}^{N_t}\sum_{i=1}^{N_x}
    \left|\nabla\!\cdot\!\mathbf{v}(\mathbf{x}_i,t_j)\right|^2 .
    \label{eq:common_physics_metrics}
\end{equation}
For PI-NOMT-NoSrc, the source contribution in $R_{PDE}$ is set to zero, consistent with that model's governing equation.
We additionally report the 99th percentile of speed magnitude and the mean absolute single-step mass change across the advection stages. This latter diagnostic reflects both residual compressibility of the learned velocity field and numerical nonconservation introduced by semi-Lagrangian interpolation. Time-averaged speed, relative source, divergence, and PDE residual maps provide complementary spatial localization of the inferred transport and physics diagnostics.

\subsection{Continuous Density Teacher Quality}

Because the frozen density teacher supplies both reconstruction targets and derivatives during transport inference, we first evaluate its fidelity to the measured tracer sequence. Supplementary Figure~\suppref{fig:teacher_quality} reports spatial reconstruction, frame-wise error, mass preservation, and voxel-density distributions, including acquisition frames withheld entirely from teacher fitting. Mean NMSE is \(0.535\%\) across all 29 frames and \(0.867\%\) on held-out frames, with 26 of 29 frames below \(1\%\). The modest late-frame increase in normalized error occurs as tracer concentration becomes more diffuse and reference energy decreases. Spatial errors, total-mass trajectories, and density distributions remain closely matched throughout the rising and decaying phases.

Because complete acquisition frames are excluded during fitting, these results additionally show that the continuous teacher reconstructs unseen acquisition-time states while preserving the measured tracer dynamics, supporting its use as the fixed continuous reference for subsequent transport inference.

\FloatBarrier

\subsection{Continuous Gaussian Density Interpolation}

The Gaussian teacher is fitted only at five integer times, so the 36 interframes directly test whether the neural field learns the continuous temporal evolution of the density field rather than memorizing separate volumes. Supplementary Figure~\suppref{fig:gaussian_teacher} shows four representative held-out interframe times for which analytic ground truth is available. Teacher NMSE is \(0.0153\%\) on the training frames and \(0.0202\%\) on the interframes. The small increase from training-frame to interframe NMSE indicates that interpolation accuracy remains close to the fitting-time reconstruction accuracy. Held-out centroid error averages \(0.104\) voxels. Thus, the teacher reconstructs the concentration field while accurately preserving the prescribed diagonal centroid trajectory at unseen times.

The teacher preserves the time-varying mass trend as well. Its mean absolute held-out mass error is \(1.291\%\), with a maximum of \(2.720\%\). This agreement supports using the teacher as dense interframe supervision for the subsequent transport model. Importantly, these analytic interframe volumes are used only for teacher evaluation and are never provided to the transport network.

\FloatBarrier

\subsection{Recovery of the Prescribed Advective Field}

\begin{figure*}[!t]
\centering
\includegraphics[width=0.9\textwidth]{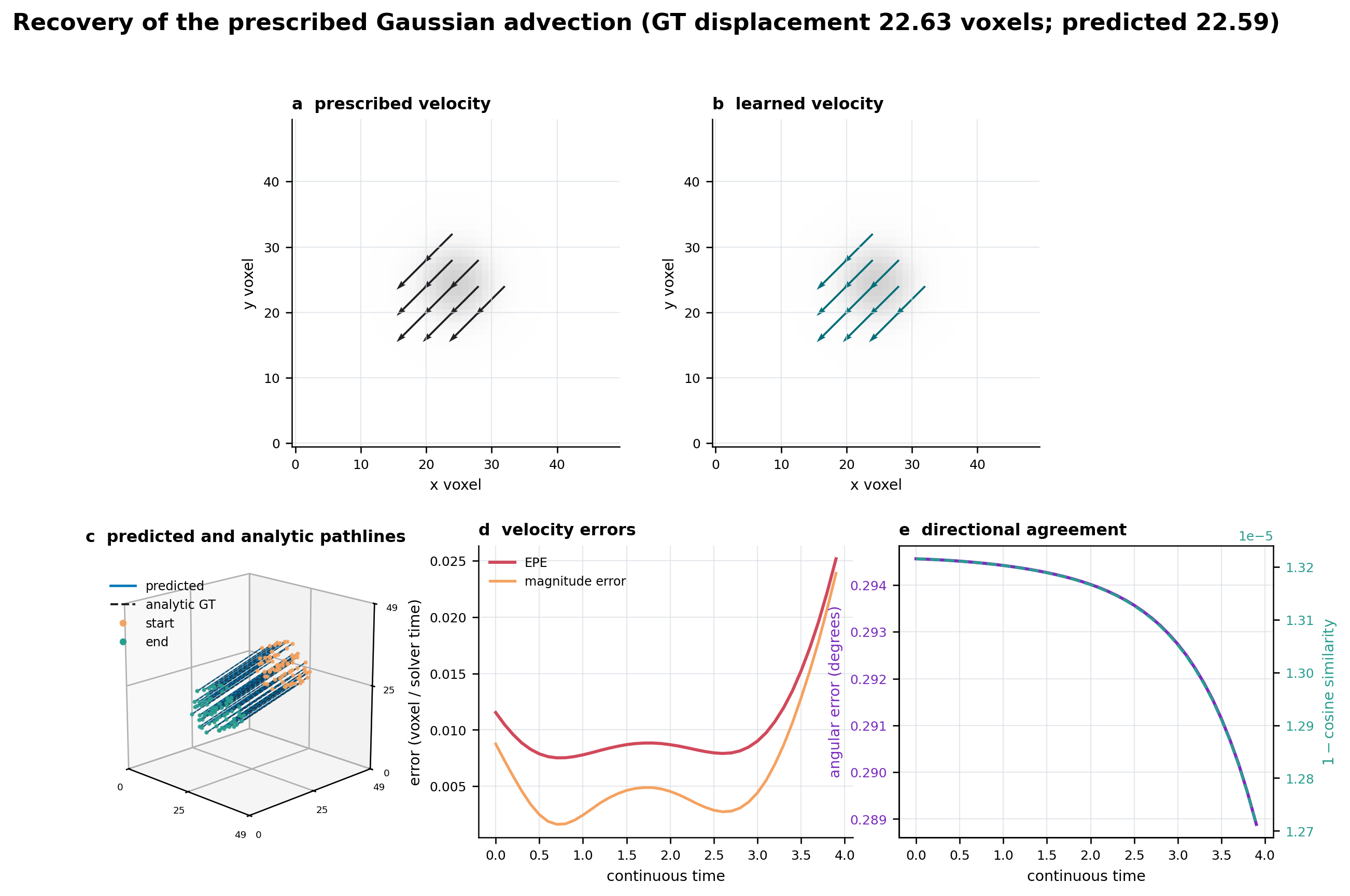}
  \caption{Recovery of the prescribed Gaussian advective field.
    The top row compares the prescribed and learned velocity fields on a
    representative central plane. The lower row compares learned pathlines
    with analytic trajectories initialized from identical seeds and shows
    magnitude and directional errors across the rollout interval.}
    \Description{Comparison of the prescribed and learned Gaussian velocity
    fields, analytic and learned trajectories from identical seeds, and
    velocity magnitude and directional errors over the rollout interval.}
    \label{fig:gaussian_field_recovery}
\end{figure*}

Figure~\ref{fig:gaussian_field_recovery} compares the learned velocity
directly with the prescribed constant translation. Across all 40 substep
times, velocity EPE and magnitude error are $0.01021$ and
$0.00605$ voxels per numerical time unit, respectively.
Relative to the ground-truth speed
$\lVert\mathbf{v}_{\mathrm{GT}}\rVert_2=\sqrt{3}\times0.8167=1.415$,
these correspond to relative errors of approximately $0.72\%$ and
$0.43\%$, respectively. Directional agreement is similarly strong:
the mean angular error is $0.293^{\circ}$, with a cosine similarity
of $0.999987$.

The Lagrangian comparison reaches the same conclusion.  Over the four acquisition intervals (16 numerical time units), the prescribed displacement is $22.632$ voxels, whereas the learned trajectories travel $22.590$ voxels on average, a relative difference of $-0.187\%$.  When analytic and learned trajectories are integrated from identical seeds, the mean Euclidean distance between their final endpoints is $0.130$ voxels, which we report as pathline EPE.

Prior urOMT work emphasized qualitative visualization of inferred transport trajectories~\cite{chen2021unbalanced}. In the present benchmark, the prescribed advective field is known analytically, allowing recovery to be assessed directly rather than from trajectory appearance alone. We therefore evaluate both Eulerian field recovery through EPE, magnitude error, angular error, and cosine similarity, and Lagrangian recovery through pathline EPE.
The quantitative results in Table~\ref{tab:gaussian_velocity}, together with the direct field and trajectory comparisons in Figure~\ref{fig:gaussian_field_recovery}, show that
the learned field closely recovers the prescribed constant translation. These results demonstrate that PI-NOMT recovers the underlying advective field itself rather than only reproducing the resulting density evolution.

\begin{table*}[!t]
\centering
\caption{Recovery of the prescribed Gaussian advective field. Field metrics
are averaged over the prescribed Gaussian evaluation support $\mathcal{S}(t)$
defined in Section~\ref{subsec:SynthGaussian} and over 40 substep times.
Velocity EPE and magnitude error are reported in voxels per numerical time unit,
angular error in degrees, and cosine similarity is dimensionless. Pathline EPE
denotes the mean Euclidean distance between the final endpoints of analytic and
learned trajectories initialized from identical seeds and is reported in voxels.}
\label{tab:gaussian_velocity}
\small
\setlength{\tabcolsep}{5pt}
\begin{tabular}{@{}lccccc@{}}
\toprule
Method
& \shortstack{Velocity\\EPE $\downarrow$}
& \shortstack{Velocity\\magnitude error $\downarrow$}
& \shortstack{Velocity\\angular error ($^\circ$) $\downarrow$}
& \shortstack{Velocity\\cosine similarity $\uparrow$}
& \shortstack{Pathline\\EPE (voxels) $\downarrow$} \\
\midrule
PI-NOMT-Cart & 0.01021 & 0.00605 & 0.293 & 0.999987 & 0.130 \\
\bottomrule
\end{tabular}
\end{table*}

\FloatBarrier

\subsection{Synthetic Source--Transport Decomposition and Rollout}

Supplementary Figure~\suppref{fig:gaussian_transport_summary} shows that the inferred relative-source field follows
the prescribed gain-then-loss sequence, while the recovered speed field remains
nearly uniform over the moving Gaussian support, as expected from the constant
analytic translation. The mean advective speed is $1.412$ voxels per numerical
time unit, whereas the diffusion-induced drift is negligible by comparison
(mean drift-to-advection ratio $0.046\%$). Thus, both the inferred source
evolution and transport field are consistent with the known synthetic
mechanism. Additional source and trajectory diagnostics, including the analytic
relative-source definition, are provided in Supplementary
Sec.~\suppref{supp:synthetic_transport}. 

To test whether the recovered fields reproduce the complete density evolution,
we initialize once at $t=0$ and recursively apply all 40 learned
source--advection--diffusion updates without resetting.
Supplementary Figure~\suppref{fig:gaussian_rollout} shows that endpoint NMSE increases from $0.133\%$ at $t=1$ to
$0.941\%$ at $t=4$, with a four-endpoint mean of $0.503\%$, when evaluated
directly against analytic ground truth rather than the density teacher.
Long-horizon mass evolution is less accurate: relative mass error increases
from $5.29\%$ to $13.56\%$ over the same interval. Thus, PI-NOMT accurately
recovers the advective field and density structure, while mass evolution remains
more sensitive to source calibration and residual numerical nonconservation.

\FloatBarrier

\subsection{Inferred Rat-Brain Transport Field}

\begin{figure*}[t]
\centering
\includegraphics[width=\textwidth]{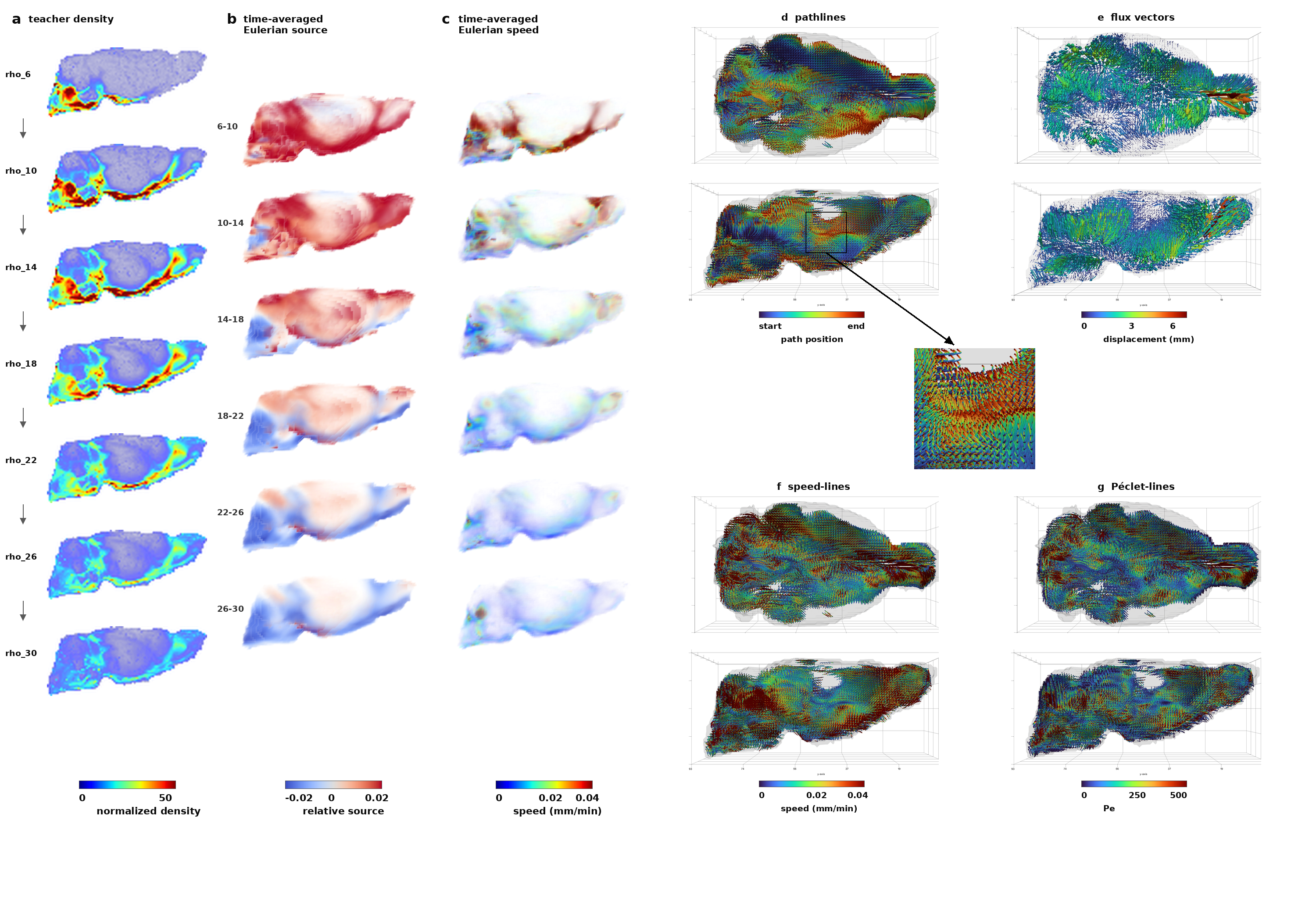}
\caption{PI-NOMT rat-brain transport summary.  Columns show density snapshots,
relative-source maps, speed maps, pathlines, flux vectors, speed-lines, and
P\'eclet-lines.  Density frames span acquisition frames 6--30.  Density, relative-source, and Eulerian speed fields are shown as translucent 3D slice renderings, using density- and scalar-dependent opacity and one shared scale per column.
The Eulerian speed maps and speed-lines use the shared physical range
0--0.04~mm/min; values above this range are clipped to the upper color limit. 
Flux-vector color denotes endpoint displacement in mm.  The Lagrangian
panels show two complementary orientations of the inferred field; the black square
in the pathline panel is connected to an enlarged detail inset between the
upper and lower panel groups.}
\Description{Seven density frames and their time-local source and speed fields
show tracer evolution from frame 6 to frame 30.  Two orientations of pathlines,
endpoint flux vectors, speed-colored trajectories, and Péclet-colored
trajectories summarize the inferred field.  A marked square in the
side pathline view connects to a magnified region of densely organized curves.}
\label{fig:main_visual_summary}
\end{figure*}

Figure~\ref{fig:main_visual_summary} summarizes the observed tracer evolution
together with the inferred relative-source and Eulerian velocity fields and
their complementary Lagrangian transport representations.

The inferred field shows a coherent, spatially structured transport pattern
rather than a uniform velocity field. High-speed regions concentrate along the
lower and posterior transport bands, while the source maps evolve from
widespread positive tracer gain at early times toward spatially separated gain
and loss regions later in the sequence. Across the anatomical mask and
evaluated time points, the mean learned Eulerian speed is
$0.00846$~mm/min ($0.141~\mu$m/s). As a complementary Lagrangian measure, the
pathline-sampled mean advective speed is $0.00839$~mm/min
($0.140~\mu$m/s). The close agreement between the Eulerian and pathline-sampled
summaries indicates a similar characteristic advective-speed scale under the
two sampling perspectives. These inferred fields and trajectory
representations provide a physically constrained, model-consistent explanation
of the observed tracer evolution; they should not be interpreted as direct
measurements of physiological fluid transport.

\FloatBarrier

\subsection{Recursive Rollout Quality}

To test whether the inferred fields generate the tracer dynamics under repeated
application, the rollout is initialized from the teacher density at frame 6 and
recursively evolved to frame 34 without resetting. The final continuous-rollout
NMSE relative to the teacher density is $26.58\%$, whereas the mean
reference-start endpoint NMSE is $0.680\%$. These metrics evaluate different
regimes: the reference-start protocol reinitializes each two-frame interval
from the teacher and therefore measures local transport accuracy, whereas the
continuous rollout is initialized only once and accumulates errors from
repeated source, advection, diffusion, and numerical interpolation updates over
the full frame 6--34 interval.

Supplementary Figure~\suppref{fig:rollout_density_error} localizes this long-horizon error accumulation
spatially. The recursive rollout preserves the dominant spatial structure of
the tracer distribution, while the error maps show where discrepancies become
progressively more pronounced over the full rollout.

\FloatBarrier

\subsection{Effect of Source Regularization}
\label{sec:SourceRegularization}

The source and velocity fields provide complementary mechanisms for explaining
an unbalanced tracer sequence: local mass gain or loss can be represented by
the relative source, whereas spatial redistribution is represented by
transport. Because both mechanisms contribute to the observed density
evolution, their inferred decomposition depends on their relative
regularization. We therefore examine
$w_{\mathrm{src}}\in\{1{,}000,5{,}000,10{,}000,20{,}000\}$ in
Eq.~\eqref{eq:total_loss}. These sensitivity runs are trained and evaluated
over frames 6--22 and thus characterize the effect of source regularization
over this restricted horizon rather than providing a direct model-selection
comparison with the full frame-6--34 experiment.

Supplementary Figure~\suppref{fig:source_weight_effect} and Table~\ref{tab:source_weight_effect} show that
increasing $w_{\mathrm{src}}$ progressively suppresses inferred source activity
but does not improve recursive prediction. The unweighted mean
$\mathcal{L}_{\mathrm{src}}$ decreases from
$3.239\times10^{-3}$ to $4.18\times10^{-4}$ across the tested range, while
continuous-rollout NMSE increases monotonically from $9.28\%$ to $20.18\%$.
Reference-start endpoint errors remain similar for the first three settings
($0.687\%$, $0.670\%$, and $0.776\%$) before increasing to $1.146\%$ at the
strongest regularization. Thus, comparable short-interval reconstruction can
coexist with substantially different source--transport decompositions, and the
small local advantage at $w_{\mathrm{src}}=5{,}000$ does not translate into
improved accumulated rollout accuracy.

The latent-field diagnostics change systematically in the opposite direction:
as source activity is suppressed, speed P99, advection-induced mass change,
PDE residual, and incompressibility all increase
(Table~\ref{tab:source_weight_effect}). These joint trends are consistent with
source--velocity compensation: stronger penalization of local mass gain and
loss shifts more of the burden of explaining the unbalanced density evolution
toward advective transport, producing faster inferred motion together with
larger physical-consistency residuals.

\newcommand{\sourceweighttablerows}{%
1{,}000 & 9.279 & 0.687 & 0.0967 & 0.00709 & 0.003239 & 0.0360 & 0.467 \\
5{,}000 & 11.247 & 0.670 & 0.1084 & 0.00937 & 0.001222 & 0.0399 & 0.568 \\
10{,}000 & 13.843 & 0.776 & 0.1243 & 0.01014 & 0.000710 & 0.0413 & 0.696 \\
20{,}000 & 20.185 & 1.146 & 0.1699 & 0.01163 & 0.000418 & 0.0433 & 0.714 \\
}

\begin{table*}[t]
\centering
\caption{Source-regularization sensitivity over frames 6--22.  Reference-start
NMSE is averaged over the eight two-frame chunk endpoints.  Mean PDE and
incompressibility are evaluated at the same 80 post-step times from 6.2 through
22.0.  Mean source regularization is the unweighted $\mathcal{L}_{\mathrm{src}}$
from Eq.~\eqref{eq:source_loss}; speed P99 is reported in mm/min, and
mean absolute advection mass change is computed over the rollout steps.}
\label{tab:source_weight_effect}
\small
\setlength{\tabcolsep}{3.8pt}
\resizebox{\textwidth}{!}{%
\begin{tabular}{@{}lrrrrrrr@{}}
\toprule
$w_{\mathrm{src}}$ & Final NMSE (\%) & Ref.-start NMSE (\%) & Mean PDE & Mean incomp. & Mean $\mathcal{L}_{\mathrm{src}}$ & Speed P99 (mm/min) & Mean $|\Delta M_{\mathrm{adv}}|$ (\%) \\
\midrule
\sourceweighttablerows
\bottomrule
\end{tabular}%
}
\end{table*}

\subsection{Comparison with urOMT on Rat-Brain Data}

The urOMT study~\cite{chen2021unbalanced} reports rat-brain endpoint NMSE
across several source-regularization values $\alpha$ in its Figure~6b.
At lower source regularization, urOMT achieves sub-percent endpoint errors,
including $0.51\%\pm0.22\%$ at $\alpha=1{,}000$ and
$0.98\%\pm0.39\%$ at $\alpha=3{,}000$. At the $\alpha=10{,}000$
value specified in its reported algorithm configuration (Table~1), the
corresponding mean NMSE is $2.58\%\pm1.28\%$. The complete set of reported
values is summarized together with PI-NOMT in
Table~\ref{tab:ratbrain_comparison}.

urOMT processes the complete image sequence through 14 successive optimization
loops. Each loop solves a new finite-dimensional transport problem for one
adjacent observed-image pair, and the terminal interpolation from the preceding
loop initializes the next. Its reported endpoint errors are therefore obtained
along a chained sequence of interval-specific optimizations, with velocity and
source variables re-estimated separately for each interval rather than generated
by a single transport field shared across the complete sequence.

For PI-NOMT, we report the closest available short-interval statistic over the
14 two-frame intervals
$6{\rightarrow}8,\ldots,32{\rightarrow}34$. Each interval is initialized from
the frozen density teacher, evolved through 10 recursive
source--advection--diffusion substeps using the same learned time-continuous
transport and source fields, and compared with the teacher at the interval
endpoint. This yields $0.680\%\pm0.679\%$ (mean and sample standard deviation).
Teacher initialization is intrinsic to PI-NOMT's two-stage formulation rather
than an additional procedure introduced for this comparison.

The resulting values should therefore be interpreted as a contextual comparison
of short-interval endpoint reconstruction rather than a strict head-to-head
evaluation under an identical protocol. urOMT chains interval-specific
solutions, whereas the PI-NOMT reference-start metric reinitializes the density
at each interval while querying the same globally learned transport model.
A matched comparison would require both methods to use the same initialization,
chaining, and evaluation procedure, which is not performed here. We restrict
the numerical comparison to urOMT because the original rOMT study
~\cite{chen2018regularized} does not report a directly comparable endpoint
value for this sequence.

The numerical source-weight value $10{,}000$ is also not directly comparable
between the two formulations. In urOMT, $\alpha$ weights the source action
within its variational objective, whereas in PI-NOMT,
$w_{\mathrm{src}}$ weights $\mathcal{L}_{\mathrm{src}}$ within the composite
objective of Eq.~\eqref{eq:total_loss}, whose reconstruction, transport, PDE,
and incompressibility terms have their own scales. Equal numerical values
therefore do not imply equal effective regularization or the same
source--velocity tradeoff.

Finally, as shown in Section~\ref{sec:SourceRegularization}, similar endpoint
errors can coexist with substantially different inferred source and transport
fields. The low-$\alpha$ urOMT results demonstrate strong short-interval
reconstruction performance, but endpoint NMSE alone does not establish
equivalence, or relative quality, of the corresponding latent transport
decompositions.

\begin{table*}[!htbp]
\centering
\caption{Rat-brain short-interval endpoint reconstruction. Representative
urOMT values are the means and standard deviations reported in the legend of
Figure~6b in Chen et al.~\cite{chen2021unbalanced} over its chained,
interval-specific optimizations. The PI-NOMT value is the mean and sample
standard deviation over 14 teacher-initialized, 10-substep reference-start
endpoint NMSE values. The comparison provides short-interval reconstruction
context under the respective inference procedures rather than a matched
optimization protocol.}
\label{tab:ratbrain_comparison}
\begin{tabular}{lc}
\toprule
Method &
Mean endpoint NMSE (\%) $\downarrow$ \\
\midrule
urOMT ($\alpha=1{,}000$) & $0.51\pm0.22$ \\
urOMT ($\alpha=3{,}000$) & $0.98\pm0.39$ \\
urOMT ($\alpha=6{,}000$) & $1.74\pm0.79$ \\
urOMT ($\alpha=10{,}000$) & $2.58\pm1.28$ \\
urOMT ($\alpha=20{,}000$) & $4.92\pm2.24$ \\
urOMT ($\alpha=50{,}000$) & $10.94\pm3.92$ \\
PI-NOMT & $0.680\pm0.679$ \\
\bottomrule
\end{tabular}
\end{table*}
\FloatBarrier

\subsection{Ablation Tradeoffs}

The ablations test both the structural components of PI-NOMT and the effect of
physics supervision within each model family. Table~\ref{tab:transport_tradeoffs}
therefore reports paired results with the PDE and incompressibility losses
switched off and on.

Physics supervision consistently improves the explicit physical-consistency
diagnostics while suppressing high-speed solutions, as summarized in
Supplementary Figure~\suppref{fig:physics_effects}. Across all five model families, enabling the physics
losses lowers the PDE residual, incompressibility, and speed P99. The effect is
especially large for PI-NOMT: the mean PDE residual decreases from 2.0644 to
0.0742, incompressibility from 0.01666 to 0.00701, and speed P99 from
0.0376 to 0.0265~mm/min. PI-NOMT is also the only model family for which
physics supervision improves both density-error protocols: final continuous
NMSE decreases from 28.75\% to 26.58\%, and mean reference-start NMSE from
0.713\% to 0.680\%. Mean absolute advection mass change increases from
0.524\% to 0.592\%, however, showing that the physics terms do not uniformly
improve every numerical diagnostic. The common-grid time series in
Supplementary Figure~\suppref{fig:common_physics_timeseries} further show how these quantities evolve over the
complete evaluation interval.

The structural ablations clarify how individual model components affect the
inferred transport solution. Removing the source branch eliminates the
model's explicit mechanism for local tracer gain and loss, leaving velocity
and the fixed diffusion term to explain the observed evolution. Relative to
physics-supervised PI-NOMT, PI-NOMT-NoSrc increases speed P99 from 0.0265 to
0.0352~mm/min and mean absolute advection mass change from 0.592\% to
0.834\%; its density errors, PDE residual, and incompressibility error also
increase. Final and reference-start NMSE rise from 26.58\% and 0.680\% to
31.97\% and 1.150\%, respectively. These changes are consistent with greater
reliance on velocity-mediated transport when the source mechanism is removed.
Physics supervision still lowers the NoSrc PDE residual from 0.7699 to 0.1734
and speed P99 from 0.0527 to 0.0352~mm/min, although its final NMSE increases
from 26.97\% to 31.97\%.

PI-NOMT-Cart instead predicts the three Cartesian velocity components directly
with a single network. In this parameterization, the expressive feature
representation acts directly on the velocity components rather than separating
magnitude from a normalized direction field. The Cartesian model exhibits more
rapid spatial variation in direction and visually less coherent pathlines in
Supplementary Figure~\suppref{fig:physical_diagnostic_maps}. This observation is considered separately from its
larger incompressibility error, since pathline curvature is not itself a
measure of divergence. Relative to PI-NOMT, its incompressibility error
increases from 0.00701 to 0.01301 and speed P99 from 0.0265 to
0.0358~mm/min. Supplementary Figure~\suppref{fig:velocity_histograms} further shows that physics supervision
contracts the high-speed tails of both NoSrc and Cart, although their P99 values
remain above that of PI-NOMT.

Physics supervision reduces the Cartesian model's PDE residual from 2.4808 to
0.0943 and incompressibility error from 0.13618 to 0.01301, but its final
continuous NMSE increases from 23.76\% to 89.00\%. Its much smaller
reference-start error of 1.048\% indicates that locally moderate errors
accumulate into an unstable long-horizon rollout, as shown directly in
Supplementary Figure~\suppref{fig:selected_nmse_curves}. These results suggest that the effect of physics
supervision depends on the velocity representation, with the
magnitude--direction factorization providing a useful inductive bias for stable
recursive transport inference beyond parameter-count differences alone.

The temporal ablations distinguish numerical resolution from recursive
supervision. PI-NOMT-2S represents the same interval with two rather than 10
source--advection--diffusion cycles, making each numerical step five times
larger. The coarser discretization substantially degrades reconstruction:
relative to PI-NOMT, final NMSE rises from 26.58\% to 39.19\%,
reference-start NMSE from 0.680\% to 0.973\%, and mean absolute advection mass
change from 0.592\% to 2.771\%. This degradation occurs despite lower PDE
residual, incompressibility, and speed, so it cannot be attributed simply to a
less physically consistent or excessively fast velocity field. Instead, the
markedly larger per-step advection mass change is consistent with reduced
temporal resolution, in which the same interval is traversed through
substantially larger numerical steps.

PI-NOMT-OneStep receives local rather than recursively coupled reconstruction
supervision. Each update starts from the teacher density, so the model is not
trained on its own predicted state and later reconstruction losses do not
propagate through earlier source and velocity decisions. Although it attains
low speed and incompressibility values, its final continuous NMSE reaches
65.05\% and its reference-start NMSE reaches 2.565\%. In the recursive model,
by contrast, each predicted density becomes the input to the next
source--advection--diffusion cycle, and losses at later intermediate states
propagate through preceding updates. Recursive supervision therefore trains the
learned fields to remain effective under repeated composition rather than only
for teacher-initialized local transitions.

\newcommand{\ablationfactorialrows}{%
\textbf{PI-NOMT$^{\dagger}$} & \textbf{26.58} & \textbf{0.680} & \textbf{0.0742} & \textbf{0.00701} & \textbf{0.0265} & \textbf{0.592} \\
PI-NOMT & 28.75 & 0.713 & 2.0644 & 0.01666 & 0.0376 & 0.524 \\
\addlinespace[2pt]
PI-NOMT-NoSrc$^{\dagger}$ & 31.97 & 1.150 & 0.1734 & 0.01243 & 0.0352 & 0.834 \\
PI-NOMT-NoSrc & 26.97 & 0.764 & 0.7699 & 0.04490 & 0.0527 & 0.918 \\
\addlinespace[2pt]
PI-NOMT-Cart$^{\dagger}$ & 89.00 & 1.048 & 0.0943 & 0.01301 & 0.0358 & 0.579 \\
PI-NOMT-Cart & 23.76 & 0.505 & 2.4808 & 0.13618 & 0.0405 & 0.729 \\
\addlinespace[2pt]
PI-NOMT-2S$^{\dagger}$ & 39.19 & 0.973 & 0.0666 & 0.00268 & 0.0209 & 2.771 \\
PI-NOMT-2S & 30.47 & 0.777 & 0.1149 & 0.00502 & 0.0281 & 3.186 \\
\addlinespace[2pt]
PI-NOMT-OneStep$^{\dagger}$ & 65.05 & 2.565 & 0.0919 & 0.00075 & 0.0110 & 0.437 \\
PI-NOMT-OneStep & 54.70 & 1.940 & 1.1414 & 0.00441 & 0.0162 & 0.646 \\
}

\begin{table*}[t]
\centering
\caption{Paired ablation of physics supervision and transport-model structure.
Continuous NMSE is measured after rolling from frame 6 to frame 34.
Reference-start NMSE is the mean two-frame chunk endpoint error when each chunk
starts from the teacher density. PDE and incompressibility are the space--time
means in Eq.~\eqref{eq:common_physics_metrics}, evaluated on the same 140 times
and all masked points for every model without density rollout. Speed P99 is
reported in mm/min, and mean absolute advection mass change is computed over
the rollout steps. Lower values indicate better reconstruction or lower
physical/numerical residuals for the corresponding error metrics; Speed P99 is
reported as a characteristic of the inferred velocity field. A superscript
$\dagger$ denotes physics supervision; unmarked rows omit the PDE and
incompressibility losses. The selected PI-NOMT formulation is marked in bold.}
\label{tab:transport_tradeoffs}
\small
\setlength{\tabcolsep}{3.5pt}
\resizebox{\textwidth}{!}{%
\begin{tabular}{@{}lrrrrrr@{}}
\toprule
Variant & Final NMSE (\%) & Ref.-start NMSE (\%) & Mean PDE & Mean incomp. & Speed P99 (mm/min) & Mean $|\Delta M_{\mathrm{adv}}|$ (\%) \\
\midrule
\ablationfactorialrows
\bottomrule
\end{tabular}%
}
\end{table*}
\FloatBarrier

Overall, the ablations show that the behavior of PI-NOMT cannot be attributed
to the addition of physics losses alone: its structural components and physics
supervision constrain different aspects of the latent transport solution.
The source branch provides an explicit mechanism for local mass variation, the
magnitude--direction factorization supplies a useful velocity-field inductive
bias, the finer 10-step temporal discretization improves the temporal resolution
of the numerical transport update, and recursive supervision trains the learned
fields to compose over time. Physics supervision further constrains the latent
solution space toward lower PDE residual, lower incompressibility, and reduced
high-speed tails, although its effect on density reconstruction depends on the
transport representation and rollout formulation. In the selected PI-NOMT
formulation, the magnitude--direction representation, source mechanism, finer
temporal discretization, and recursive supervision act jointly with physics
supervision to balance density reconstruction, physical consistency, and
recursive rollout stability.

\subsection{Cross-Subject Consistency}
\label{sec:cross_subject}

Beyond the detailed C1217 experiment, we independently applied the complete
two-stage PI-NOMT pipeline to eight additional control-rat sequences.
Table~\ref{tab:cohort_summary} summarizes reconstruction, velocity, and
physical-consistency metrics across all nine subjects. The mean reference-start
endpoint NMSE remains below one percent
($0.801\%\pm0.167\%$), with a range of $0.532$--$1.090\%$.
For the uninterrupted frame-6--34 rollout, the mean same-time NMSE averaged
over the rollout is $11.773\%\pm2.606\%$ across subjects. Thus, local
reconstruction remains consistently accurate, while accumulated error over the
full recursive rollout exhibits moderate subject-to-subject variation.

\begin{table*}[t]
\centering
\caption{Cross-subject consistency over nine control-rat DCE-MRI sequences.
Values are the mean $\pm$ sample standard deviation across subjects. The
continuous-rollout entry first averages same-time teacher NMSE over frames
6--34 within each subject. PDE and incompressibility entries are post-training
space--time means obtained by directly querying the neural fields. Eulerian
speed summarizes each subject's full-interval time-averaged velocity field,
whereas pathline-sampled mean advective speed pools the local advective
magnitude over all mask-inside trajectory-time samples of the corresponding
integrated trajectories.}
\label{tab:cohort_summary}
\begin{tabular}{lcc}
\toprule
Metric & Mean $\pm$ SD & Subject range \\
\midrule
Reference-start endpoint NMSE (\%) & $0.801 \pm 0.167$ & $0.532$--$1.090$ \\
Continuous-rollout NMSE (\%) & $11.773 \pm 2.606$ & $6.555$--$15.298$ \\
Mean Eulerian speed (mm/min) & $0.00738 \pm 0.00111$ & $0.00533$--$0.00930$ \\
Pathline-sampled mean advective speed (mm/min) & $0.00711 \pm 0.00108$ & $0.00508$--$0.00876$ \\
Speed P99 (mm/min) & $0.02968 \pm 0.00592$ & $0.02235$--$0.03989$ \\
PDE residual MSE & $0.07010 \pm 0.02977$ & $0.02997$--$0.13078$ \\
Incompressibility MSE & $0.00629 \pm 0.00124$ & $0.00453$--$0.00776$ \\
\bottomrule
\end{tabular}
\end{table*}

The inferred velocity scale is also relatively consistent across the cohort:
the mean Eulerian speed is
$0.00738 \pm 0.00111$~mm/min
($0.1230 \pm 0.0185~\mu\mathrm{m/s}$), while the pathline-sampled mean
advective speed is
$0.00711 \pm 0.00108$~mm/min
($0.1186 \pm 0.0180~\mu\mathrm{m/s}$).
The close agreement between these cohort-level summaries indicates a similar
characteristic advective-speed scale under Eulerian and trajectory-based
sampling. These estimates are also consistent in physical scale with
solute-transport speeds previously reported from rOMT-based Lagrangian analysis
of rat DCE-MRI~\cite{chen2022cerebral}.

\begin{figure*}[p]
\centering
\includegraphics[width=\textwidth]{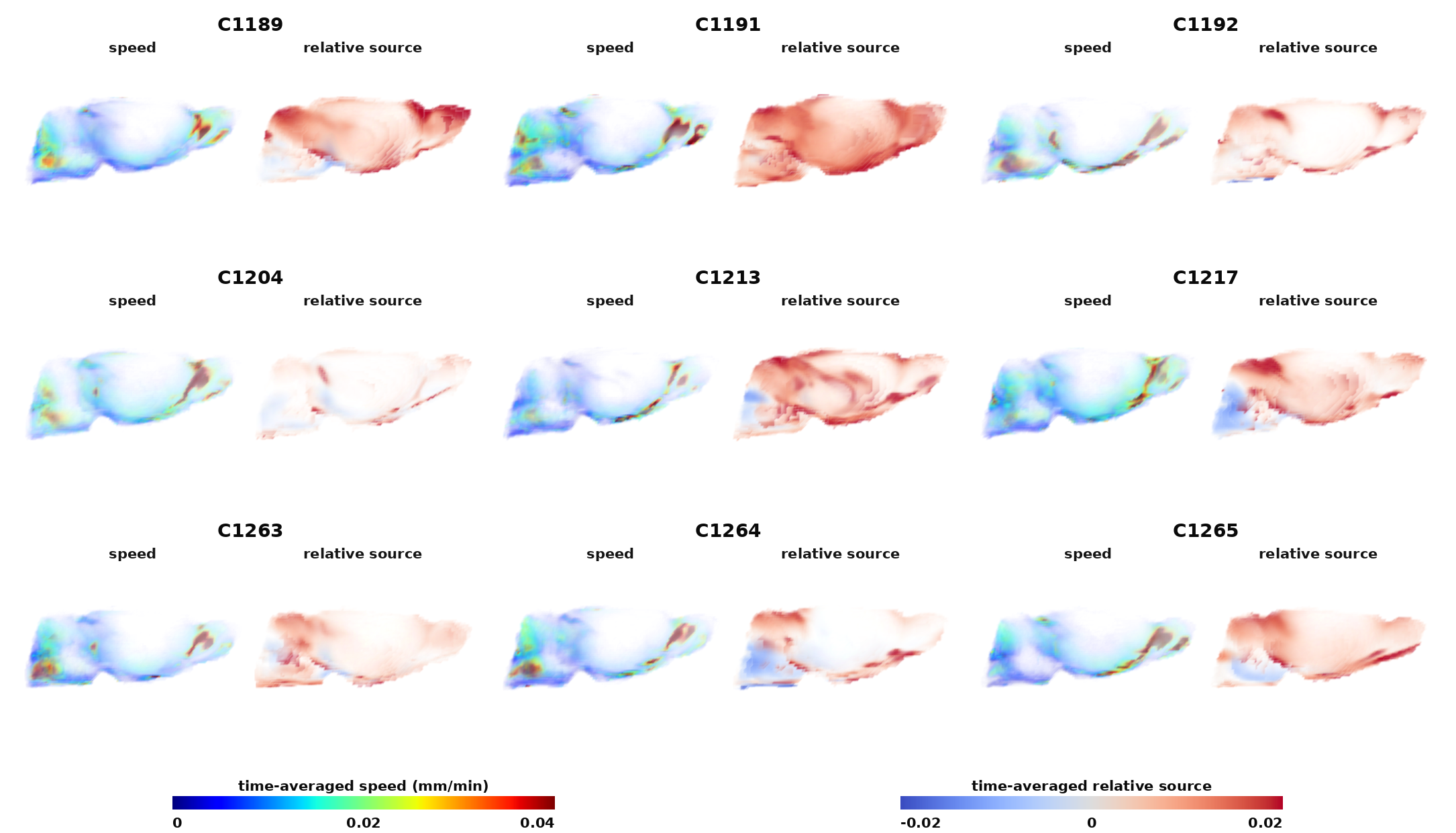}
\caption{Cross-subject Eulerian transport comparison across all nine
control-rat sequences. Each subject is shown with its time-averaged speed and
relative-source fields over frames 6--34. Speed and source panels use shared
scales of 0--0.04~mm/min and $[-0.02,0.02]$, respectively.}
\Description{A three-by-three arrangement shows time-averaged Eulerian speed
and relative-source maps for all nine control-rat subjects under shared color
and opacity scales.}
\label{fig:cohort_eulerian}
\end{figure*}

Figure~\ref{fig:cohort_eulerian} provides the corresponding spatial
comparison across all nine subjects under shared visualization scales.
The inferred fields consistently exhibit localized high-speed structure and
spatially organized positive- and negative-relative-source regions
(sources and sinks), while retaining subject-specific differences in their
extent and intensity. Corresponding pathline and speed-line visualizations in
Supplementary Figures~\suppref{fig:cohort_appendix_pathlines}
and~\suppref{fig:cohort_appendix_speedlines} provide complementary Lagrangian views and
likewise show coherent directional organization across the illustrated
subjects.

Together with the cohort-level reconstruction and physical-consistency metrics
in Table~\ref{tab:cohort_summary}, these results support cross-subject
consistency in the inferred latent-field characteristics within this control
cohort. They do not imply that the subject-specific fields are anatomically
identical or constitute direct measurements of physiological fluid velocity.
Additional Eulerian and trajectory visualizations are provided in
Supplementary Appendix~\suppref{supp:cross_subject_visualizations}.

\FloatBarrier

\section{Discussion}
\label{sec:discussion}

The experimental evaluation first establishes the validity of the continuous
density teacher. The teacher reproduces the measured tracer dynamics with low
reconstruction error on held-out frames while preserving the overall mass
evolution and density distribution. These results indicate that the teacher
provides an accurate continuous representation of the observed tracer dynamics,
including at held-out acquisition times. Consequently, the transport model can
be supervised at arbitrary spatial and temporal locations while obtaining the
spatial and temporal derivatives required by the governing transport equations.
This decoupled formulation separates density reconstruction from transport
inference, providing a fixed continuous reference throughout transport
learning. At the same time, transport inference is conditional on this learned
density representation: inaccuracies in its temporal interpolation or spatial
and temporal derivatives can propagate into the rollout targets and physics
residuals.

The rollout ablations further clarify the role of differentiable physical rollout in latent transport inference. The teacher-reset one-step model, which is trained only on local transitions, produces substantially poorer recursive rollout because it is never exposed to the accumulation of its own prediction errors. Similarly, reducing the rollout resolution from ten to two substeps degrades long-horizon density reconstruction despite maintaining relatively low local PDE residuals. 
Together, these observations show that, within the present formulation, neither local transition fitting nor pointwise physical consistency alone is sufficient for stable recursive transport inference.
Instead, recursive rollout supervision imposes multi-substep temporal consistency within each sampled window by requiring the learned velocity and source fields to repeatedly generate the intermediate teacher states. The uninterrupted long-horizon rollout then evaluates how well the shared fields compose beyond these training windows.

The physics ablation studies further demonstrate that governing equations influence not only reconstruction accuracy but also the characteristics of the recovered latent transport fields. Models with comparable density reconstruction can infer substantially different velocity and source fields, with corresponding differences in PDE residual, incompressibility, and high-speed behavior. Incorporating physics supervision consistently reduces the explicit PDE and incompressibility residuals and contracts high-speed tails across all evaluated model families.
However, the Cartesian parameterization reveals that improved physical consistency alone does not guarantee improved reconstruction. As shown in Table~\ref{tab:transport_tradeoffs}, adding physics constraints substantially improves the physical metrics for PI-NOMT-Cart while markedly degrading reconstruction accuracy, showing that the effect of physics supervision depends strongly on the velocity representation. These observations suggest that the primary role of physics supervision is not simply to improve reconstruction performance, but to reduce the admissible latent transport solutions to those that remain consistent with the governing transport dynamics. Consequently, the governing equations act as structural priors that restrict the space of admissible latent transport fields.

While physics supervision constrains the admissible transport solutions, the
comparison between velocity parameterizations further suggests that an
appropriate inductive bias is equally important for physics-informed transport
inference. As illustrated quantitatively in
Table~\ref{tab:transport_tradeoffs} and visually in
Supplementary Figure~\suppref{fig:physics_effects}, PI-NOMT is the only evaluated model family
for which improved physical consistency translates into improvements in both
local density reconstruction and long-horizon transport prediction. 
The recovered fields exhibit localized variations in velocity magnitude together with coherent directional organization, consistent with the intended inductive bias of the magnitude--direction factorization. By representing magnitude and direction separately, this parameterization allows their spatial structure to be modeled through distinct branches, allowing localized transport-speed variations to be represented separately from smoother directional organization.
Together with the physics ablation results, these findings suggest that recovering latent transport fields is not determined solely by the governing equations; physics supervision is most effective when paired with a representation whose inductive bias is aligned with the structure of the underlying transport process.

The source-weight sensitivity study further illustrates the inherent ambiguity of inverse transport inference. Increasing the source penalty progressively suppresses inferred mass creation and destruction, but the optimization compensates through larger transport velocities, increased physical residuals, and poorer long-horizon rollout accuracy. Interestingly, multiple source-weight settings achieve comparable local density reconstruction despite producing substantially different recursive transport behavior. In the present formulation, this decomposition is additionally conditional on the fixed diffusion coefficient \(D\); allowing \(D\) to vary spatially or temporally, or inferring it jointly, would introduce a richer but also less constrained latent decomposition. These observations are consistent with the transport--mass-change tradeoff formalized by Wasserstein--Fisher--Rao and dynamic unbalanced optimal transport~\cite{chizat2018interpolating,chizat2018unbalanced}, rather than implying that PI-NOMT computes either metric exactly. More broadly, they demonstrate that local density reconstruction alone does not uniquely determine the underlying latent transport mechanism, highlighting the importance of combining optimal transport regularization, recursive rollout, and physics supervision to constrain the space of admissible latent transport solutions.

The cross-subject evaluation shows that these findings are not confined to the
detailed C1217 experiment.  Across nine control-rat sequences, PI-NOMT achieves a mean local endpoint error below 1\%, together with a consistent physical speed scale across subjects and low post-training PDE and incompressibility residuals. The recurring organization of Eulerian fields and trajectories supports cross-subject consistency of the inferred latent transport characteristics while retaining subject-specific variation.

The synthetic benchmark provides a unique validation of latent transport-field recovery because the prescribed latent transport field is known. The recovered velocity field closely matches the prescribed constant diagonal transport, exhibiting low velocity EPE, angular, magnitude, and pathline endpoint errors while accurately reproducing the gain-then-loss density evolution. Furthermore, the continuous density teacher reconstructs the unseen interframe sphere trajectory, allowing the learned transport field to be evaluated continuously rather than only at the observed frames. These results demonstrate that PI-NOMT recovers the prescribed advective transport field itself, rather than merely reconstructing its endpoint densities.

Importantly, the recovered velocity and source fields are recursively integrated to generate future tracer evolution beyond local density reconstruction. Successful long-horizon rollout therefore provides complementary evidence that the learned transport fields form a dynamically consistent model of the evolving density rather than simply interpolating between observations. The synthetic experiment also highlights that visually more complex trajectories are not necessarily indicative of more accurate transport recovery. For the prescribed constant translational field, the correct solution consists of nearly straight trajectories, demonstrating that direct quantitative evaluation of latent transport fields complements qualitative visualization and provides a more objective assessment of transport recovery. The long-horizon mass errors observed in the synthetic experiment may also partly reflect the numerical properties of the rollout. The backward semi-Lagrangian advection step provides a stable and differentiable rollout but is not strictly mass conservative, so improved differentiable conservative transport operators are a natural direction for extending the present formulation. Conservative PINN formulations that enforce flux continuity provide one related strategy for learning conservation-law solutions~\cite{jagtap2020conservative}, although they do not make the present semi-Lagrangian update conservative.

The rat-brain experiments demonstrate that PI-NOMT recovers the same class of
biologically meaningful transport quantities previously obtained with
transport-based approaches~\cite{chen2018regularized,chen2021unbalanced,
koundal2020optimal,koundal2024divergent}, while doing so through a fundamentally
different physics-informed latent-state inference framework. The inferred
relative-source field exhibits a clear temporal transition, with predominantly
positive values during the early phase consistent with local tracer gain,
followed by increasingly negative values during the later phase consistent with
local tracer loss. This evolution follows the measured total tracer-mass trend
and is qualitatively consistent with gain-and-loss behavior previously reported
for this dataset~\cite{chen2021unbalanced}. In addition, the inferred velocity
fields exhibit localized high-speed regions together with coherent transport
trajectories, providing a model-based description of the inferred transport
process throughout the imaging interval.

These latent fields provide a physically constrained, model-consistent
representation of the observed tracer evolution. The inferred velocity and
source fields are jointly constrained by the observed data, unbalanced optimal
transport regularization, recursive physical rollout, and the governing
transport equations. PI-NOMT therefore complements existing transport-based
methods by providing an alternative inference paradigm for estimating these
established transport quantities rather than introducing new biological
observables.

These experimental observations suggest that PI-NOMT provides a framework for recovering physically constrained latent transport mechanisms from observed density evolution. The combination of continuous teacher supervision, recursive differentiable rollout, unbalanced optimal transport regularization, and continuous physics supervision progressively constrains the inferred fields while maintaining consistency with the observed tracer dynamics.
Importantly, this reformulation does not redefine the biological quantities recovered from tracer imaging; rather, it changes the computational methodology by which they are estimated, enabling continuous neural representations, recursive differentiable rollout, and validation against known transport fields in synthetic experiments.
More broadly, PI-NOMT introduces a physics-informed latent-state inference framework for recovering continuous latent transport fields from time-resolved scalar density or concentration observations. Although demonstrated here on brain tracer transport, the same latent-state inference perspective may extend to other dynamic imaging and transport problems in which evolving scalar fields are observable while the underlying transport and source mechanisms remain hidden.

\paragraph{Limitations}
The principal limitation is the inherent non-identifiability of latent
transport from tracer density alone. PI-NOMT reduces this ambiguity through
physics supervision, unbalanced optimal transport regularization, recursive
rollout, and architectural inductive biases, but these constraints do not
establish uniqueness of the inferred velocity and source fields. The recovered
rat-brain fields should therefore be interpreted as model-consistent,
physically constrained explanations of the observed tracer evolution rather
than direct measurements of physiological fluid transport.

The present biological evaluation is also limited to nine control-rat
sequences acquired under a common protocol. Broader validation across
independent cohorts, acquisition settings, and physiological or disease
conditions will be required to establish the generalizability of the inferred
transport characteristics.

\section{Conclusion}
\label{sec:conclusion}

PI-NOMT formulates the recovery of continuous transport velocity and source fields from sparse time-resolved scalar observations as a physics-informed latent-state inference problem. It replaces interval-specific voxelwise optimization with continuous neural transport fields trained through recursive differentiable transport and governing-equation supervision. Synthetic experiments show direct recovery of known advective dynamics, while rat-brain DCE-MRI experiments demonstrate accurate short-interval reconstruction, physically constrained latent fields, and consistent transport characteristics across subjects. The ablation studies further show that governing-equation supervision is most effective when combined with an appropriate latent-field representation and recursive transport formulation, highlighting the interaction among physical constraints, inductive bias, and long-horizon consistency. These results establish PI-NOMT as a computational framework for recovering physically interpretable transport quantities from dynamic imaging data, while the inferred fields remain model-based estimates rather than direct physiological measurements. More broadly, the framework provides a general latent-state perspective for inverse transport problems in which hidden dynamics must be inferred from observable evolving scalar fields.

\begin{acks}
This research was supported by the NVIDIA Academic Grant Program using two NVIDIA RTX 6000 Ada GPUs. The authors gratefully acknowledge NVIDIA Corporation for providing the hardware that enabled the computational experiments reported in this work.
\end{acks}

\section*{Dedication}
This work is dedicated to the memory of Allen Tannenbaum, whose pioneering contributions to optimal transport, medical imaging, and mathematical data science continue to inspire new generations of researchers. His mentorship, scientific vision, and intellectual influence are deeply reflected in the ideas that motivated this work.

\FloatBarrier


\clearpage
\appendix
\setcounter{figure}{0}
\section*{Supplementary Appendix}
This appendix contains the supporting figures referenced as
Supplementary Figures~\ref{fig:teacher_quality}--\ref{fig:cohort_appendix_speedlines}
in the main manuscript.

\renewcommand{\thefigure}{S\arabic{figure}}

\renewcommand{\thesection}{\Alph{section}}

\section{Background on Optimal Transport and Scientific Machine Learning}
\label{app:background}

\subsection{Dynamic Optimal Mass Transport}

Optimal Mass Transport (OMT) provides a mathematical framework for describing
the movement of mass between probability distributions. Originally introduced
by Monge~\cite{supp-bib:monge1781memoire} and later generalized by
Kantorovich~\cite{supp-bib:kantorovich1942translocation}, OMT seeks the least-cost
transformation that transports an initial mass distribution to a target
distribution.

A major development was the dynamic formulation of Benamou and
Brenier~\cite{supp-bib:benamou2000computational}, which recasts OMT as a fluid transport
problem. Transport is described through a time-varying density field
$\rho(x,t)$ and velocity field $\mathbf{v}(x,t)$ satisfying

\begin{equation}
\frac{\partial \rho}{\partial t}
+
\nabla \cdot (\rho \mathbf{v})
=
0.
\end{equation}

The corresponding optimization problem is

\begin{equation}
\min_{\rho,\mathbf{v}}
\int_0^1 \int_\Omega
\rho(x,t)\,|\mathbf{v}(x,t)|^2\,dx\,dt,
\end{equation}

subject to the continuity equation and fixed endpoint densities. This dynamic
viewpoint connects optimal transport with fluid mechanics and provides a
principled basis for recovering velocity fields from evolving density
observations.

\subsection{Unbalanced and Regularized Optimal Transport}

Classical OMT assumes purely advective dynamics and exact mass conservation,
which can be restrictive in biological systems involving diffusion, tracer
dispersion, production, clearance, or exchange between compartments. Dynamic
unbalanced optimal transport extends the continuity-equation formulation by
jointly representing spatial transport and mass variation. The
Wasserstein--Fisher--Rao construction combines transport with Fisher--Rao mass
change~\cite{supp-bib:chizat2018interpolating}, while related dynamic and Kantorovich
formulations formalize transport, creation, and destruction between
nonnegative measures~\cite{supp-bib:chizat2018unbalanced}. Eulerian algorithms and
variational splitting schemes provide numerical treatments of unbalanced
transport and general advection--reaction--diffusion problems
\cite{supp-bib:lombardi2015eulerian,supp-bib:gallouet2019splitting}.

Regularized OMT incorporates diffusion into the transport process
\cite{supp-bib:chen2018regularized}, while urOMT additionally introduces source and sink
terms~\cite{supp-bib:chen2021unbalanced}. These developments provide the conceptual
transport--source setting from which PI-NOMT draws; PI-NOMT does not compute an
exact unbalanced-transport geodesic or reproduce these numerical algorithms.

\subsection{Physics-Informed and Operator Learning}

Scientific Machine Learning incorporates governing equations, conservation laws,
and other physical constraints into data-driven models. Physics-Informed Neural
Networks (PINNs), for example, penalize PDE residuals at collocation points and
have been applied to forward and inverse problems involving fluid dynamics,
diffusion, reaction--diffusion systems, and transport
\cite{supp-bib:raissi2019physics}.

A broader class of scientific machine-learning approaches includes neural
operators, DeepONets, and Fourier Neural Operators, which learn mappings between
function spaces and have been widely used for surrogate modeling and simulation
acceleration~\cite{supp-bib:lu2021learning,supp-bib:li2021fourier}. These methods are conceptually
related through their use of neural representations for physical systems, but
PI-NOMT addresses a different inference problem: recovering latent transport
fields whose recursive physical evolution explains an observed density
sequence.

\subsection{Neural-Field Architectures and Density-Teacher Training}
\label{supp:network_details}

\paragraph{Density teacher.}
The continuous density teacher
$\mathcal{N}_{\rho}$ is implemented as a residual fully connected network
with four residual blocks, hidden width 256, and tanh activations. A Chebyshev
feature map augments the spatiotemporal coordinates with low-order polynomial
basis functions before they are processed by the residual network.

Let
\[
\mathcal{D}_{obs}
=
\{(\mathbf{x}_j,t_i,\rho^{obs}(\mathbf{x}_j,t_i))\}
\]
denote the observed tracer samples. The density network is initially optimized
using the supervised loss
\begin{equation}
    \mathcal{L}_{\rho}
    =
    \frac{1}{|\mathcal{B}|}
    \sum_{(\mathbf{x}_j,t_i)\in\mathcal{B}}
    \left(
    \rho_\phi(\mathbf{x}_j,t_i)
    -
    \rho^{obs}(\mathbf{x}_j,t_i)
    \right)^2 ,
    \label{eq:supp_density_loss}
\end{equation}
with mini-batches $\mathcal{B}\subset\mathcal{D}_{obs}$.
For the final 1,000 training epochs, MSE is replaced by MAE.

Residual-based adaptive weighting follows the residual-based attention (RBA)
principle~\cite{supp-bib:anagnostopoulos2024residual}; related weighting is used in
MR-AIV~\cite{supp-bib:toscano26}, although PI-NOMT does not use its time-dependent
normalization. The weighting emphasizes samples with larger current fitting
residuals during density-teacher training. After training, the density-network
parameters are frozen and remain fixed throughout transport-field inference.

\paragraph{Transport fields.}
The source, magnitude, and direction fields are represented by separate
residual fully connected networks. Each network contains eight residual blocks,
has hidden width 128, and uses tanh activations. The magnitude branch receives
the Chebyshev feature representation, whereas the source and direction
networks receive the spatiotemporal coordinates directly. This implements the
main-text inductive bias that transport-speed magnitude may exhibit more
localized spatial variation, while directional structure is modeled more
smoothly.

The full set of trainable transport parameters is
\[
\boldsymbol{\theta}
=
\{\boldsymbol{\theta}_s,
  \boldsymbol{\theta}_m,
  \boldsymbol{\theta}_d\}.
\]
Unless otherwise stated, PI-NOMT denotes the magnitude--direction
parameterization defined in the main manuscript. PI-NOMT-Cart uses the same
transport-learning framework, training objective, and rollout procedure, but
replaces the magnitude--direction factorization with a network that predicts
the three Cartesian velocity components directly.

\subsection{Differentiable Rollout: Numerical Formulation and Implementation}
\label{supp:rollout_implementation}

PI-NOMT evolves the rolled-out density through a differentiable operator
splitting of the advection--diffusion--source equation
\begin{equation}
    \rho_t+\nabla\cdot(\rho\mathbf{v})
    =
    D\nabla^2\rho+\rho r .
    \label{eq:supp_rollout_pde}
\end{equation}
For an interval $[t_a,t_b]$ divided into $K$ substeps of size
$\Delta t=(t_b-t_a)/K$, each substep separates the dynamics into source,
advection, and diffusion:
\begin{align}
    \partial_t \rho &= \rho r,
    \label{eq:supp_split_source}\\
    \partial_t \rho + \nabla\cdot(\rho\mathbf{v}) &= 0,
    \label{eq:supp_split_advection}\\
    \partial_t \rho &= D\nabla^2\rho .
    \label{eq:supp_split_diffusion}
\end{align}
Related unbalanced-transport splitting schemes treat advection, reaction, and
diffusion within a unified variational framework
\cite{supp-bib:gallouet2019splitting}. The differentiable update used here is distinct:
it combines an explicit relative-source step, semi-Lagrangian advection, and
implicit diffusion.

\paragraph{Source update.}
Let $\hat{\rho}^{k}(\mathbf{x})$ denote the current rolled-out density at
substep $k$, with source and velocity fields $r^k(\mathbf{x})$ and
$\mathbf{v}^k(\mathbf{x})$. The source equation is discretized pointwise as
\begin{align}
    \frac{
    \hat{\rho}^{k,\mathrm{src}}(\mathbf{x})
    -
    \hat{\rho}^{k}(\mathbf{x})
    }{\Delta t}
    &=
    \hat{\rho}^{k}(\mathbf{x})r^k(\mathbf{x}),
    \\
    \hat{\rho}^{k,\mathrm{src}}(\mathbf{x})
    &=
    \left(1+\Delta t\,r^k(\mathbf{x})\right)
    \hat{\rho}^{k}(\mathbf{x}).
    \label{eq:supp_source_step}
\end{align}
This is a first-order discretization of the local reaction equation. For a
nonnegative input density, the update remains nonnegative provided
\begin{equation}
    1+\Delta t\,r^k(\mathbf{x})\geq0.
\end{equation}
Source regularization and sufficiently small rollout substeps are used to
promote this condition and limit unstable multiplicative factors.

\paragraph{Semi-Lagrangian advection.}
Advection is approximated using a first-order characteristic backtrace. For
each output location $\mathbf{x}$,
\begin{align}
    \mathbf{x}_d
    &=
    \mathbf{x}
    -
    \Delta t\,\mathbf{v}^k(\mathbf{x}),
    \\
    \hat{\rho}^{k,\mathrm{adv}}(\mathbf{x})
    &=
    \mathcal{I}
    \!\left[
    \hat{\rho}^{k,\mathrm{src}}
    \right](\mathbf{x}_d),
    \label{eq:supp_advection_step}
\end{align}
where $\mathcal{I}$ denotes differentiable trilinear sampling.

The semi-Lagrangian update follows the characteristic form
\begin{equation}
    \rho_t+\mathbf{v}\cdot\nabla\rho=0.
\end{equation}
For the conservative transport equation,
\begin{equation}
    \rho_t+\nabla\cdot(\rho\mathbf{v})=0,
\end{equation}
expansion of the flux divergence gives
\begin{equation}
    \rho_t
    +
    \mathbf{v}\cdot\nabla\rho
    +
    \rho\nabla\cdot\mathbf{v}
    =
    0.
\end{equation}
Thus, the characteristic and conservative forms coincide for a divergence-free
velocity field. Motivated by the approximate incompressibility assumption used
in brain-fluid transport modeling~\cite{supp-bib:toscano26}, PI-NOMT adopts an
approximately incompressible formulation. Rather than applying an explicit
Jacobian correction to the semi-Lagrangian update, deviations from
incompressibility are controlled through the divergence penalty in the
physics-informed objective. This same modeling assumption is used in the
continuous physics residual, maintaining consistency between the rollout and
physics supervision.

\paragraph{Implicit diffusion.}
The diffusion subproblem is discretized by backward Euler,
\begin{align}
    \frac{
    \hat{\rho}^{k,\mathrm{diff}}
    -
    \hat{\rho}^{k,\mathrm{adv}}
    }{\Delta t}
    &=
    D\nabla_h^2
    \hat{\rho}^{k,\mathrm{diff}},
    \\
    \left(
    I-\Delta t D\nabla_h^2
    \right)
    \hat{\rho}^{k,\mathrm{diff}}
    &=
    \hat{\rho}^{k,\mathrm{adv}} .
    \label{eq:supp_diffusion_step}
\end{align}

The diffused density becomes the input to the next source--advection--diffusion
cycle,
\begin{equation}
    \hat{\rho}^{k+1}
    =
    \hat{\rho}^{k,\mathrm{diff}},
    \label{eq:supp_rollout_state}
\end{equation}
so the rollout is recursive and is not reset to the teacher between substeps.

\paragraph{Implicit-solve implementation and differentiation.}
The backward-Euler diffusion system is solved on the rollout grid using
conjugate gradients. The finite-difference Laplacian uses Neumann-like
replicate padding at the cropped tensor boundary, and the linear operator is
\begin{equation}
    A z
    =
    z-\Delta t D\nabla_h^2 z .
\end{equation}

Backpropagation through the implicit solve is implemented using the
corresponding adjoint linear system. If
\begin{equation}
    g
    =
    \frac{\partial\mathcal{L}}
    {\partial\hat{\rho}^{k,\mathrm{diff}}},
\end{equation}
the backward pass solves
\begin{equation}
    A q=g
\end{equation}
and returns $q$ as the gradient with respect to the diffusion-step input.
Under the implemented boundary treatment, the discrete diffusion operator is
self-adjoint, so $A^T=A$ and the backward solve uses the same operator.
Because $D$ is treated as a fixed hyperparameter, gradients propagate through
the density input but not through $D$.

\subsection{Random-Window Training and Gradient Implementation}
\label{supp:training_details}

For completeness, the transport networks are optimized using the objective
defined in the main manuscript,
\begin{equation}
    \mathcal{L}
    =
    w_{final}\mathcal{L}_{final}
    + w_{inter}\mathcal{L}_{inter}
    + w_{omt}\mathcal{L}_{omt}
    + w_{src}\mathcal{L}_{src}
    + w_{phys}\mathcal{L}_{phys}
    + w_{inc}\mathcal{L}_{inc}.
    \label{eq:supp_total_loss}
\end{equation}
Here, $\mathcal{L}_{final}$ and $\mathcal{L}_{inter}$ supervise the endpoint
and intermediate states of the recursive rollout;
$\mathcal{L}_{omt}$ and $\mathcal{L}_{src}$ regularize transport and source
activity along the predicted density trajectory; and
$\mathcal{L}_{phys}$ and $\mathcal{L}_{inc}$ impose the governing-equation
residual and approximate incompressibility, respectively. The explicit loss
definitions are given in Sec.~3.5 of the main manuscript. The details below
describe how these terms are evaluated during random-window training and how
their gradients are organized computationally.

Transport-field training uses randomly sampled temporal rollout windows to
combine trajectory-level reconstruction with local physics supervision. At
each optimization step, a temporal window is sampled continuously over the
valid training interval, providing coverage of both observed and interframe
times. Windows that intersect a temporal boundary are truncated to the
available interval. The resulting rollout times are
\begin{equation}
    t_k=t_a+k\Delta t,
    \qquad
    k=0,\ldots,K,
    \qquad
    t_b=t_a+K\Delta t,
    \label{eq:supp_window_times}
\end{equation}
where $t_a$ denotes the start of the valid portion of the sampled window and
$\Delta t$ is the fixed rollout substep size. The nominal rollout contains
$K=10$ substeps; for boundary-truncated windows, $K$ is reduced to the largest
number of complete substeps that remain within the valid temporal domain.

The frozen density teacher provides the continuous target sequence
\begin{equation}
    \rho_\phi^k(\cdot)
    =
    \rho_\phi(\cdot,t_k),
    \qquad
    k=0,\ldots,K .
    \label{eq:supp_teacher_targets}
\end{equation}
Because $\rho_\phi$ is continuous in time, the target times
$t_0,\ldots,t_K$ need not coincide with measured DCE-MRI acquisition frames.
The teacher therefore supplies both the rollout initialization and
reconstruction targets at interframe locations where direct measurements are
unavailable.

At each substep, the neural source and velocity fields are evaluated as
\begin{equation}
    r^k(\cdot)
    =
    r_{\boldsymbol{\theta}}(\cdot,t_k),
    \qquad
    \mathbf{v}^k(\cdot)
    =
    \mathbf{v}_{\boldsymbol{\theta}}(\cdot,t_k).
    \label{eq:supp_neural_substep_fields}
\end{equation}
The sampled window is initialized once from the teacher density,
\begin{equation}
    \hat{\rho}^{0}=\rho_\phi^0,
\end{equation}
and is then evolved recursively using the source--advection--diffusion operators
defined in Supplementary Sec.~\ref{supp:rollout_implementation}:
\begin{align}
    \hat{\rho}^{k,\mathrm{src}}
    &=
    \left(1+\Delta t\,r^k\right)\hat{\rho}^{k},
    \\
    \hat{\rho}^{k,\mathrm{adv}}
    &=
    \mathcal{A}_{\Delta t}
    \left(
    \hat{\rho}^{k,\mathrm{src}};
    \mathbf{v}^k
    \right),
    \\
    \hat{\rho}^{k,\mathrm{diff}}
    &=
    \mathcal{D}_{\Delta t}
    \left(
    \hat{\rho}^{k,\mathrm{adv}};
    D
    \right),
    \\
    \hat{\rho}^{k+1}
    &=
    \hat{\rho}^{k,\mathrm{diff}},
    \qquad
    k=0,\ldots,K-1 .
    \label{eq:supp_training_rollout}
\end{align}
Thus, a sampled window produces the complete recursively generated sequence
$\hat{\rho}^{1},\ldots,\hat{\rho}^{K}$ rather than only an endpoint
prediction. The rolled density is not reset to the teacher between substeps.

The endpoint reconstruction loss compares $\hat{\rho}^{K}$ with the teacher
target at the end of the window, whereas the intermediate reconstruction loss
supervises the preceding rolled-out states
$\hat{\rho}^{1},\ldots,\hat{\rho}^{K-1}$. Both losses are therefore computed
from the same recursive trajectory. When $K=1$, no intermediate rollout state
exists and $\mathcal{L}_{inter}=0$.

The OMT-style kinetic and source regularizers are evaluated along this same
predicted density trajectory. As defined in the main manuscript,
$\operatorname{sg}(\cdot)$ denotes a stop-gradient operation applied to the
rolled-density weights in $\mathcal{L}_{omt}$ and $\mathcal{L}_{src}$. The
density therefore acts as a weighting measure for velocity and source activity
without allowing these regularizers to alter earlier rollout states indirectly
through their density weights. Gradients from these terms act directly on the
corresponding transport fields.

In contrast, the recursive rollout state itself is not detached between
substeps. Gradients from $\mathcal{L}_{final}$ and
$\mathcal{L}_{inter}$ therefore propagate backward through the complete
source--advection--diffusion chain within the sampled window, including earlier
evaluations of
$\mathbf{v}_{\boldsymbol{\theta}}$ and
$r_{\boldsymbol{\theta}}$. This recurrent gradient path couples transport-field
estimation across multiple substeps rather than reducing training to a
collection of independent local transitions.

The physics and incompressibility objectives are evaluated at the individual
sampled substep times using the frozen density teacher and the continuous
transport fields. In particular, the teacher supplies the density derivatives
required for the governing-equation residual, while the learned velocity and
source networks receive the resulting gradients. The teacher parameters remain
frozen throughout transport-field training.

For memory efficiency, gradients from the physics and incompressibility losses
are evaluated and accumulated at the active substeps separately from the
recurrent reconstruction graph. This alters only the organization of gradient
computation, not the corresponding terms in the optimization objective. It
preserves the complete recurrent gradient path for reconstruction while
avoiding the need to retain the full higher-order physics graph throughout the
entire rollout window.
\section{Density-Teacher Validation}

\begin{figure*}[!htbp]
\centering
\includegraphics[width=\textwidth]{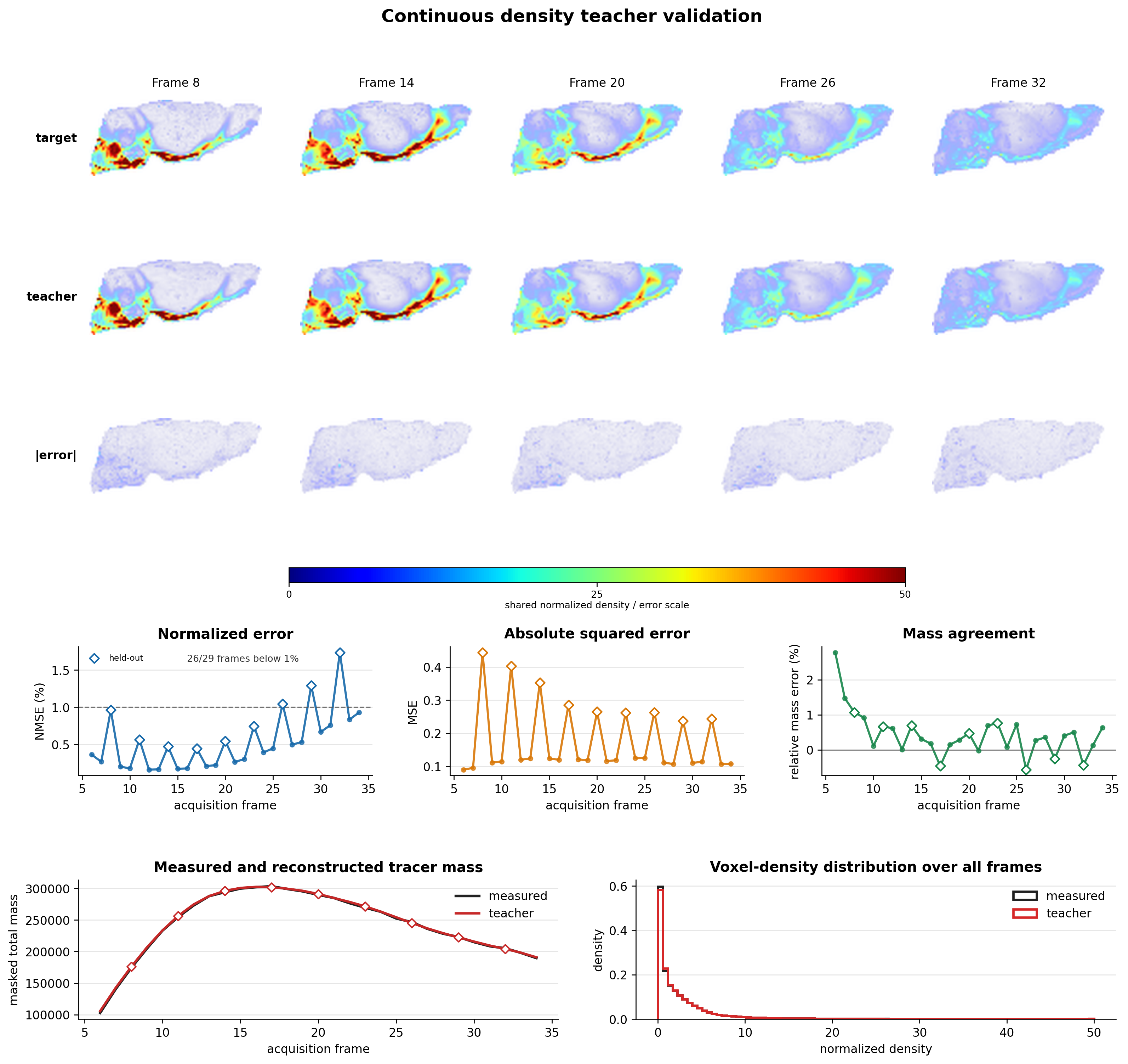}
\caption{Continuous density teacher validation. The top grid compares measured
target density, neural teacher density, and absolute error at five representative held-out validation frames using a shared normalized density/error scale. The lower panels report frame-wise NMSE, MSE, relative mass error, measured and teacher total mass, and the voxel-density distributions across all frames. Open diamond markers identify acquisition frames excluded from teacher fitting.}
\Description{Five held-out density volumes are reconstructed with faint spatial errors.  Below them, NMSE is below one percent at 26 of 29 frames, MSE remains small in absolute units, the teacher follows the measured mass trajectory, and the measured and reconstructed voxel-density distributions nearly overlap.}
\label{fig:teacher_quality}
\end{figure*}

\begin{figure*}[!htbp]
\centering
\includegraphics[width=\textwidth]{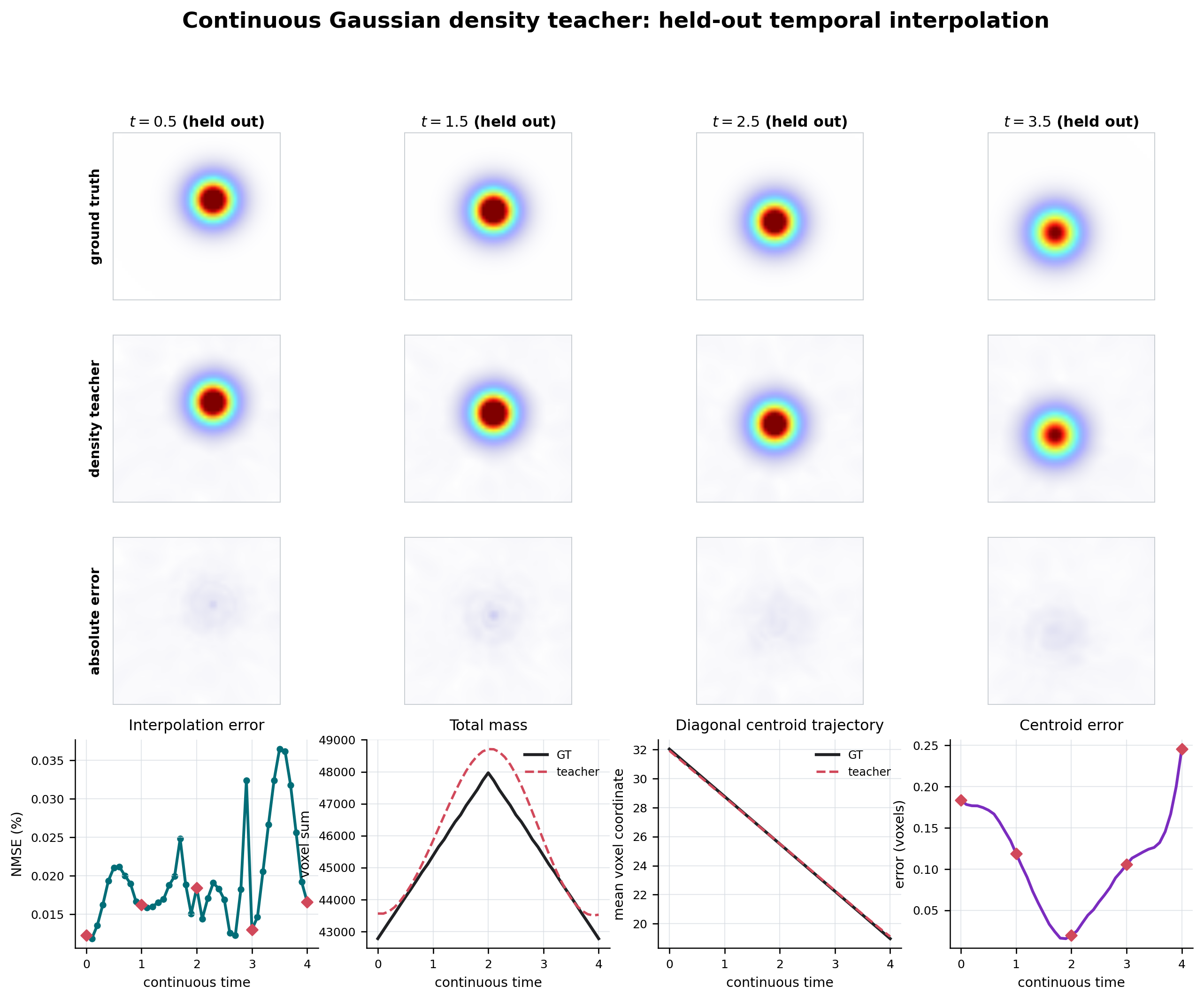}
\caption{Continuous density interpolation on the Gaussian benchmark.  The
upper rows compare analytic ground truth, density-teacher prediction, and
absolute error at four held-out interframe times.
The lower panels show dense-time NMSE, total mass,
the centroid trajectory projected onto the prescribed diagonal, and centroid
error.  Diamonds mark the five integer times used to fit the teacher.}
\Description{Ground-truth and teacher Gaussian spheres nearly coincide at four
held-out times.  Error images are faint on the shared density scale, while the
metric panels show low interpolation error, matching mass evolution, and an
accurately recovered diagonal centroid trajectory.}
\label{fig:gaussian_teacher}
\end{figure*}

\FloatBarrier

\section{Additional Synthetic Results}

\begin{figure*}[!htbp]
\centering
\includegraphics[width=\textwidth]{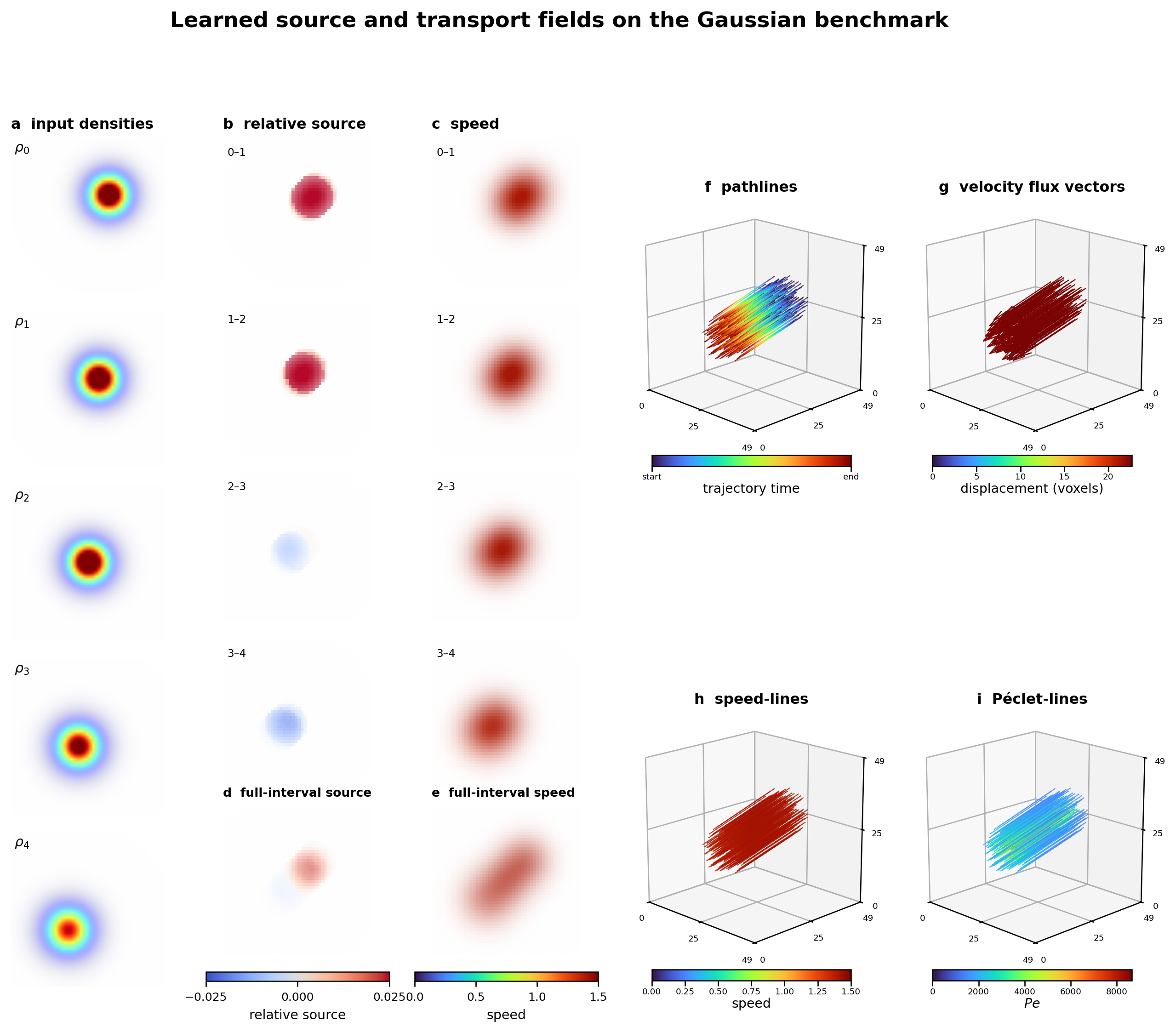}
\caption{Learned source and transport fields on the Gaussian benchmark.
Columns show the five input densities, interval-wise relative-source and speed
maps, full-interval averages, pathlines, endpoint flux vectors, speed-lines,
and P\'eclet-lines.  Trajectory geometry incorporates the diffusive drift 
$-D\nabla\log\rho$, whereas trajectory speed coloring represents the magnitude of the raw advective velocity $\left\|\mathbf{v}_{\boldsymbol{\theta}}\right\|_2$.  Eulerian maps use tracer-dependent opacity and one common
scale per quantity.  The four trajectory panels include their corresponding
trajectory-time, displacement, speed, and P\'eclet color scales.}
\Description{The Gaussian sphere moves diagonally while its inferred source is
positive in the first two intervals and negative in the final two.  Velocity
maps and trajectory renderings show a smooth, coherent, nearly constant
transport field.}
\label{fig:gaussian_transport_summary}
\end{figure*}

\begin{figure*}[!htbp]
\centering
\includegraphics[width=\textwidth]{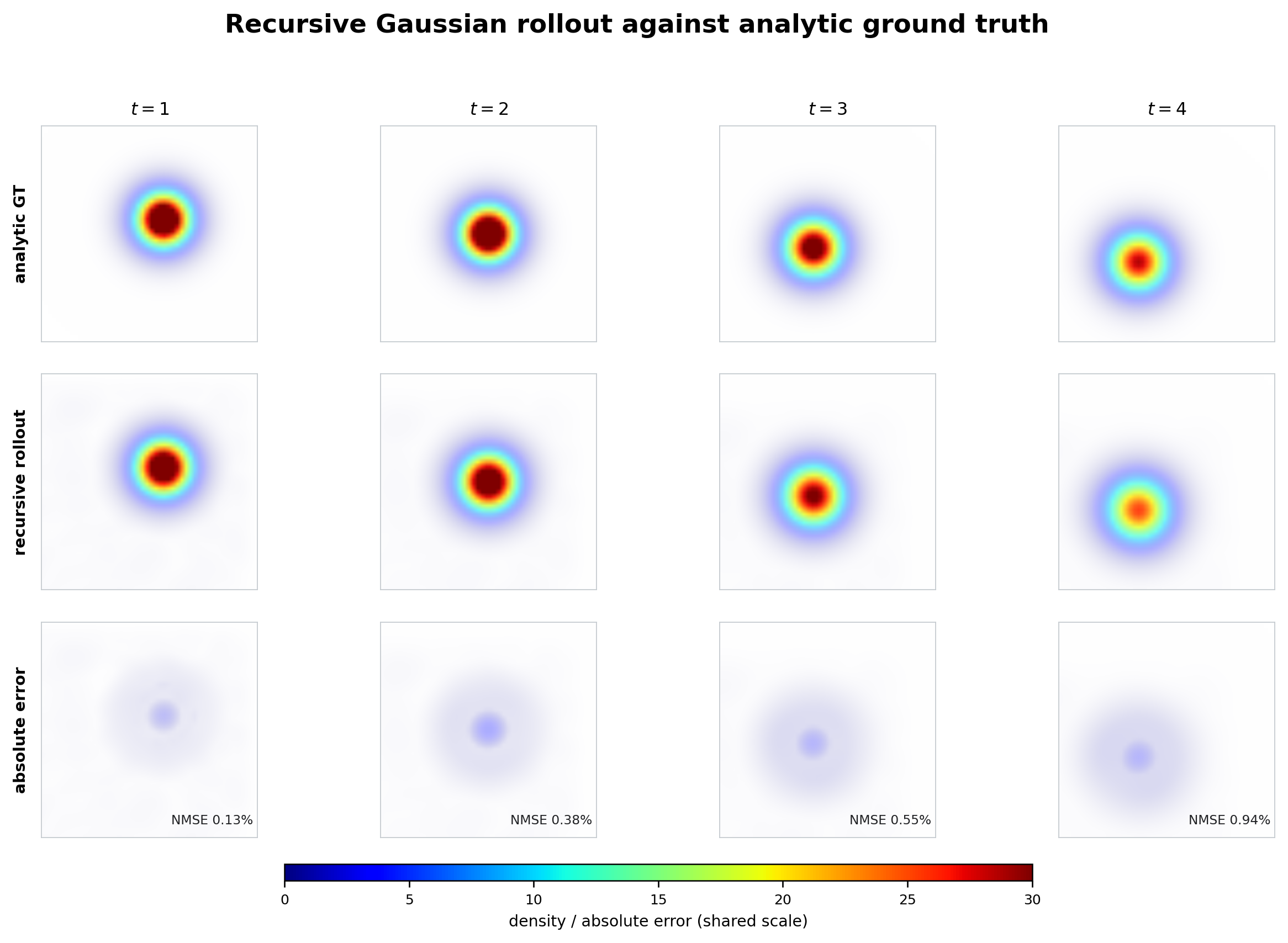}
\caption{Recursive Gaussian rollout against analytic ground truth.  The model
is initialized once at $t=0$ and recursively evolved for 40 substeps.  Analytic
density, rollout density, and absolute error share the single 0--30 color scale
shown beneath all three rows. NMSE values at each integer endpoint are reported below the corresponding error maps.}
\Description{Analytic and recursively predicted Gaussian spheres remain
spatially aligned through time.  Absolute-error maps remain faint on the shared
density scale, while NMSE rises gradually from 0.13 to 0.94 percent.}
\label{fig:gaussian_rollout}
\end{figure*}

\FloatBarrier

\subsection{Synthetic Source and Transport Diagnostics}
\label{supp:synthetic_transport}

Supplementary Figure~\ref{fig:gaussian_transport_summary} summarizes the complete inferred transport on the
Gaussian benchmark. The relative-source field is positive over intervals
$0$--$1$ and $1$--$2$, and negative over intervals $2$--$3$ and $3$--$4$,
matching the prescribed gain-then-loss sequence. The speed maps remain
spatially smooth and nearly constant over the moving tracer support, as
expected from the prescribed constant translation.

For the analytic amplitude schedule $s(t)$ used to generate the benchmark, the
corresponding relative source within the source support is
\begin{equation}
    r_{\mathrm{GT}}(t)
    =
    \frac{s'(t)}{4s(t)},
\end{equation}
where the factor $4$ accounts for the four numerical time units represented by
each acquisition-frame interval. The resulting ground-truth relative source
lies within $[-0.025,0.025]$. The source maps in Supplementary Figure~\ref{fig:gaussian_transport_summary} use
this prescribed range rather than a range fitted to the learned extrema.

The pathline, flux-vector, speed-line, and P\'eclet-line visualizations provide
complementary representations of the same coherent diagonal transport.
Trajectory geometry in Supplementary Figure~\ref{fig:gaussian_transport_summary} is generated using the
diffusion-augmented drift
$\mathbf{v}_{\boldsymbol{\theta}}-D\nabla\log\hat{\rho}$, whereas speed coloring
represents the raw advective magnitude
$\|\mathbf{v}_{\boldsymbol{\theta}}\|_2$.
Within the prescribed Gaussian support, the mean advective-speed magnitude is
$1.412$ voxels per numerical time unit, while the mean diffusion-induced drift
magnitude is $6.50\times10^{-4}$. The mean drift-to-advection ratio is therefore
$0.046\%$, and its 99th percentile is approximately $0.068\%$, explaining the
negligible visible effect of diffusion on the trajectory geometry in this
benchmark.

\FloatBarrier
\section{Rat-Brain Rollout Diagnostics}

\begin{figure*}[!htbp]
\centering
\includegraphics[width=\textwidth]{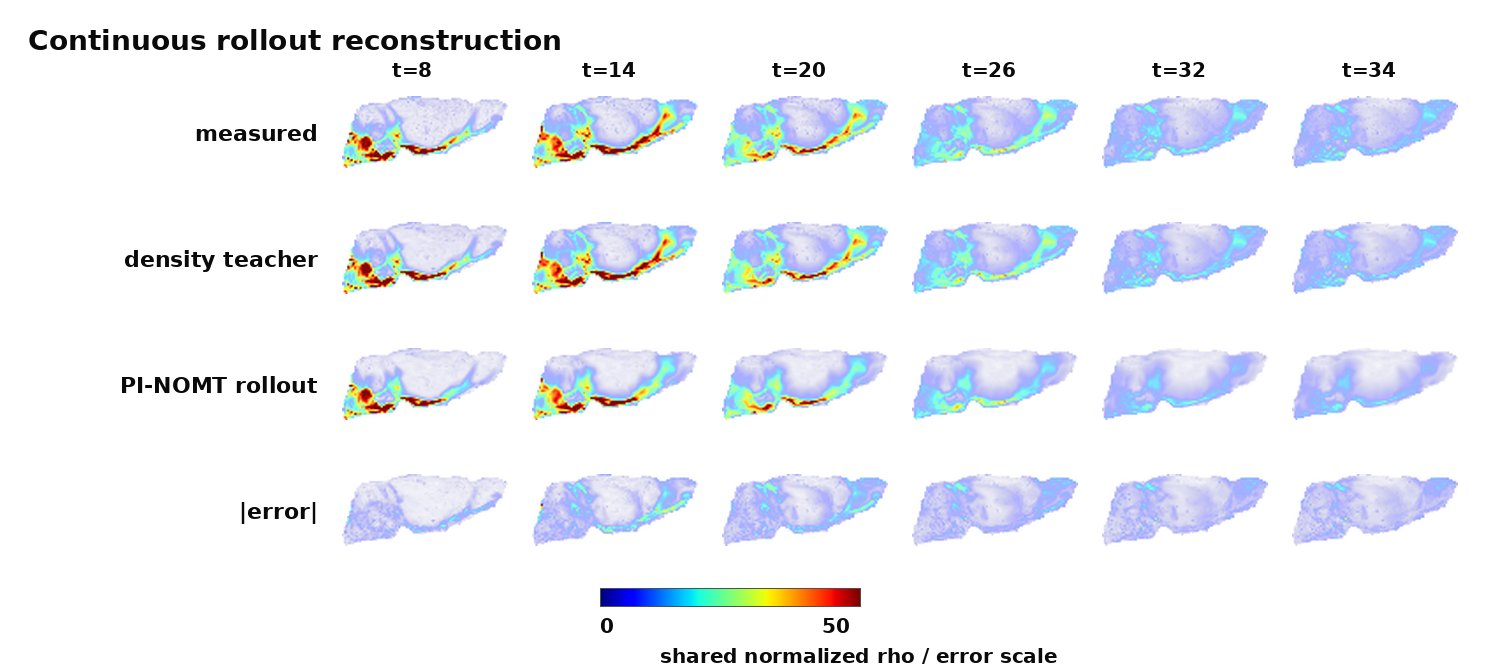}
\caption{Spatial rollout reconstruction for PI-NOMT.  The rollout is
initialized at frame 6 and recursively evolved to the selected future frames. Density and absolute-error panels use the same colormap and numerical scale, allowing error magnitude to be interpreted directly relative to the density signal.}
\Description{Grid showing measured density, density-teacher projections,
PI-NOMT rollout projections, and absolute rollout--teacher error at selected
acquisition frames.}
\label{fig:rollout_density_error}
\end{figure*}

\FloatBarrier

\section{Source-Regularization Diagnostics}

\begin{figure*}[!htbp]
\centering
\includegraphics[width=\textwidth]{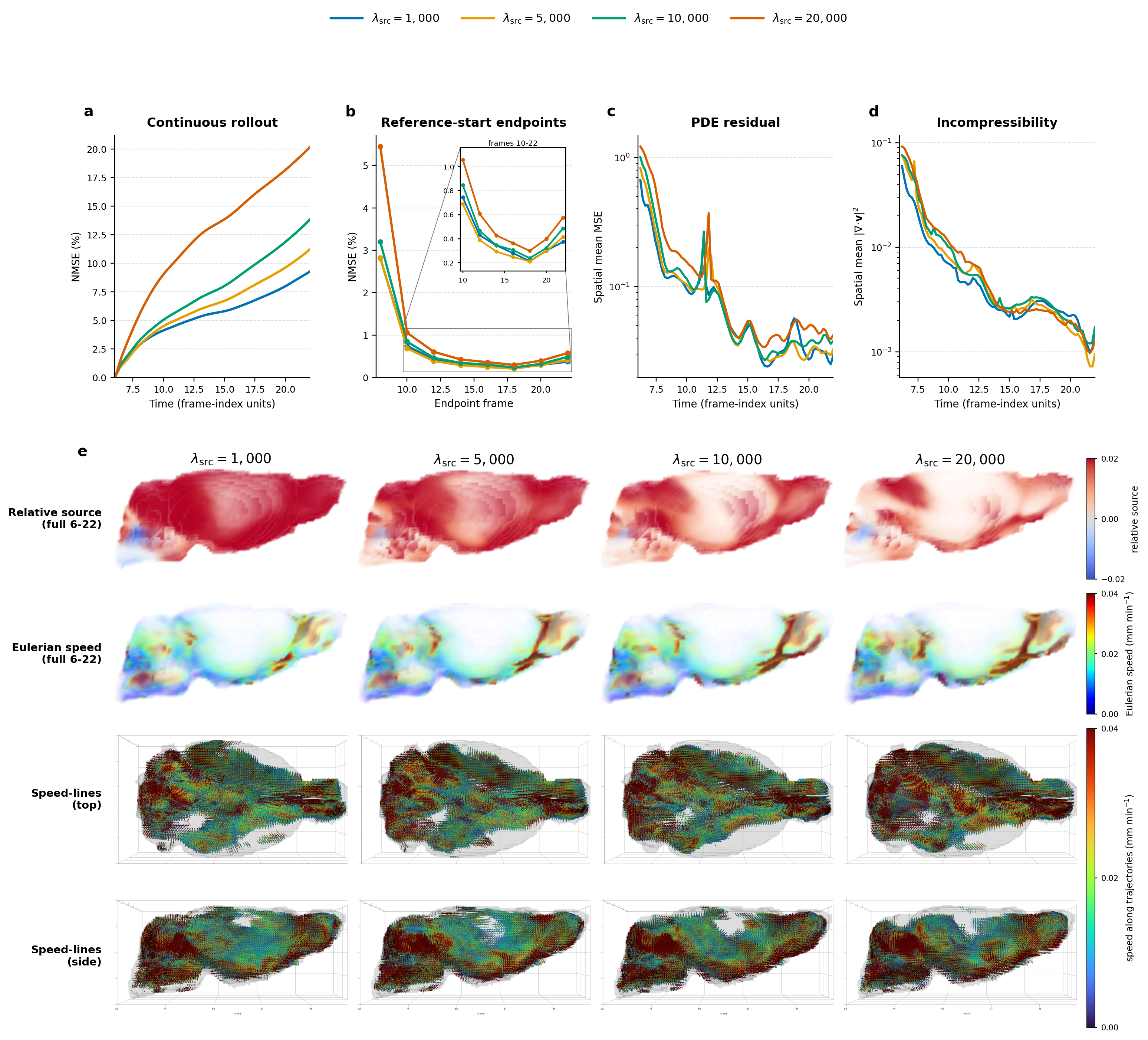}
\caption{Effect of source regularization over frames 6--22.  (a) Continuous
rollout NMSE from a single initialization at frame 6.  (b) Reference-start
two-frame endpoint NMSE, with an inset enlarging frames 10--22.  (c--d) Spatial
mean PDE residual and incompressibility evaluated on the common 80-time grid.
(e) Full-interval relative-source and Eulerian-speed maps followed by top and side
views of speed-lines for four source weights.  Source maps share the range
$[-0.02,0.02]$, while Eulerian speed and speed-lines share the physical
range $[0,0.04]$~mm/min, with values above this range clipped to the upper color limit.
The Eulerian opacity depends jointly on tracer density and the displayed scalar.}
\Description{Four temporal plots and a four-by-four visual comparison show the
effect of increasing source regularization.  Higher source weight weakens the
relative-source field, increases continuous rollout error, and produces stronger
high-speed transport together with larger PDE and incompressibility residuals.}
\label{fig:source_weight_effect}
\end{figure*}

\FloatBarrier

\section{Detailed Ablation Diagnostics}

\begin{figure*}[!htbp]
\centering
\includegraphics[width=\textwidth]{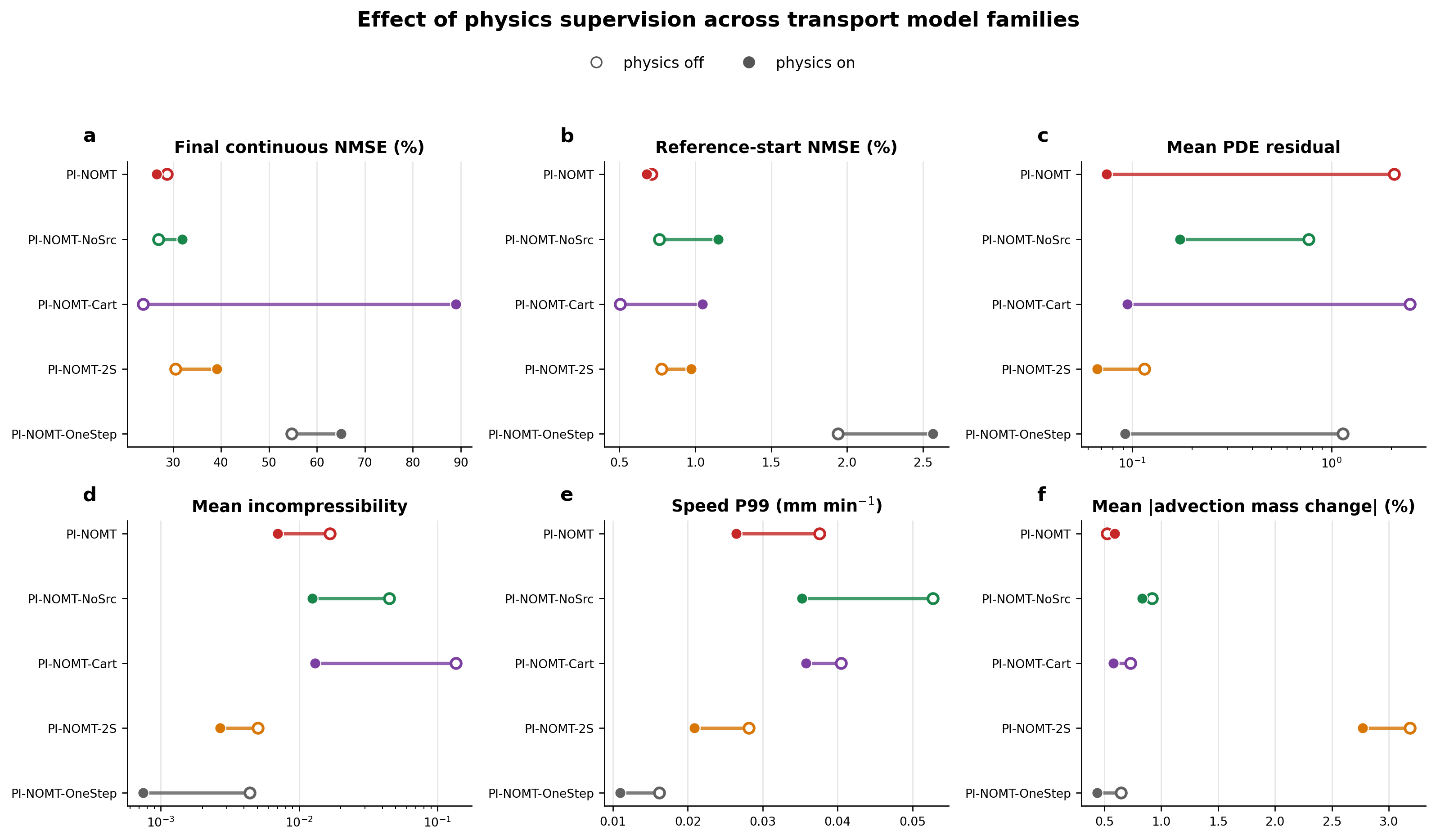}
\caption{Effect of physics supervision within five transport-model families.
Open and filled markers denote physics off and on, respectively; connected
markers share the same architecture and rollout design.  Panels report
(a) final continuous-rollout NMSE, (b) mean reference-start endpoint NMSE,
(c) common-grid PDE residual, (d) common-grid incompressibility, (e) speed P99,
and (f) mean absolute single-step advection mass change.  Lower values indicate better reconstruction or lower physical/numerical residuals for the corresponding error metrics; speed P99 is reported as a characteristic of the inferred velocity field. 
Physics supervision improves the field-consistency diagnostics in every family,
whereas PI-NOMT is the only family that also improves both density-error metrics.}
\Description{Six paired-dot panels compare physics-off and physics-on versions
of PI-NOMT, NoSrc, Cart, 2S, and OneStep.  Physics lowers PDE residual,
incompressibility, and speed P99 throughout, while density errors improve only
for PI-NOMT.}
\label{fig:physics_effects}
\end{figure*}

\begin{figure*}[!htbp]
\centering
\includegraphics[width=\textwidth]{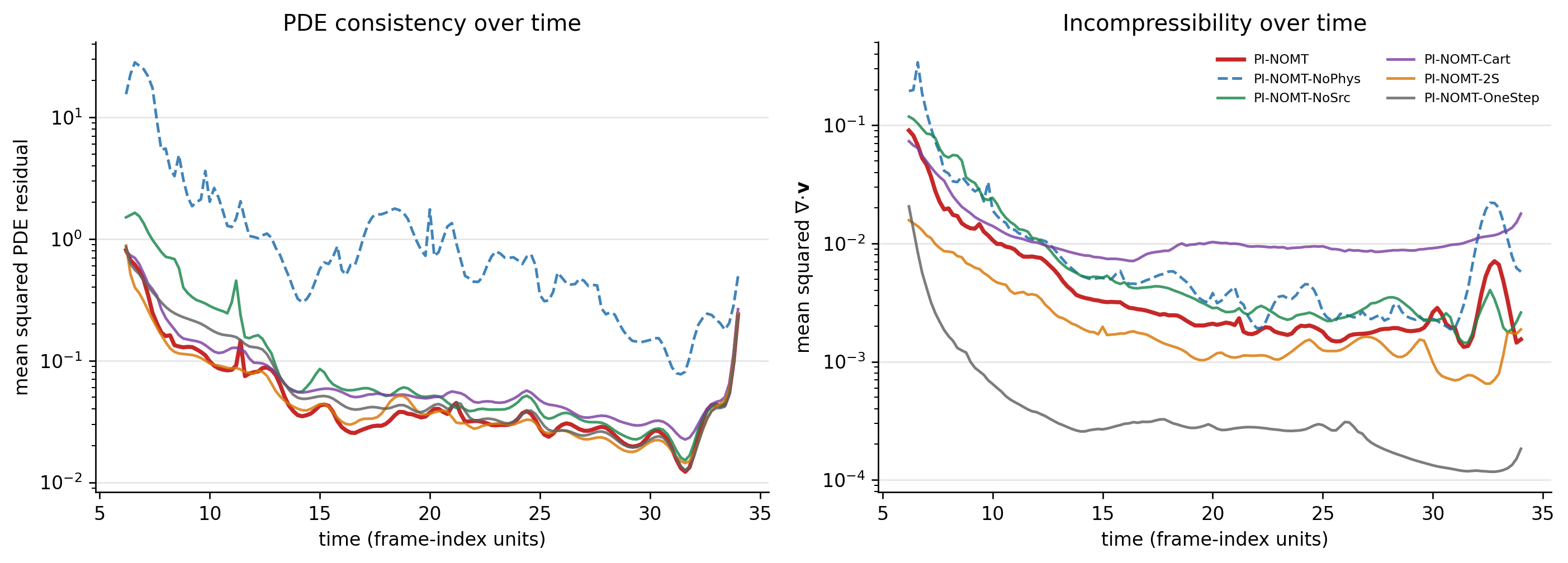}
\caption{Post-training physical consistency on the common evaluation grid.
Curves report the spatial mean squared PDE residual and incompressibility at
each of 140 shared post-step times.  The evaluation directly queries the frozen
density, velocity, and source networks and does not use rolled-out density
states.  Except for PI-NOMT-NoPhys, all displayed variants use physics
supervision.}
\Description{Two logarithmic line plots compare six transport models over
continuous evaluation time.  PI-NOMT maintains low PDE residual throughout
the interval.  PI-NOMT-NoPhys remains substantially higher, while the
physics-enabled structural and rollout ablations are regularized toward the
PI-NOMT range.}
\label{fig:common_physics_timeseries}
\end{figure*}

\begin{figure*}[!htbp]
\centering
\includegraphics[width=\textwidth]{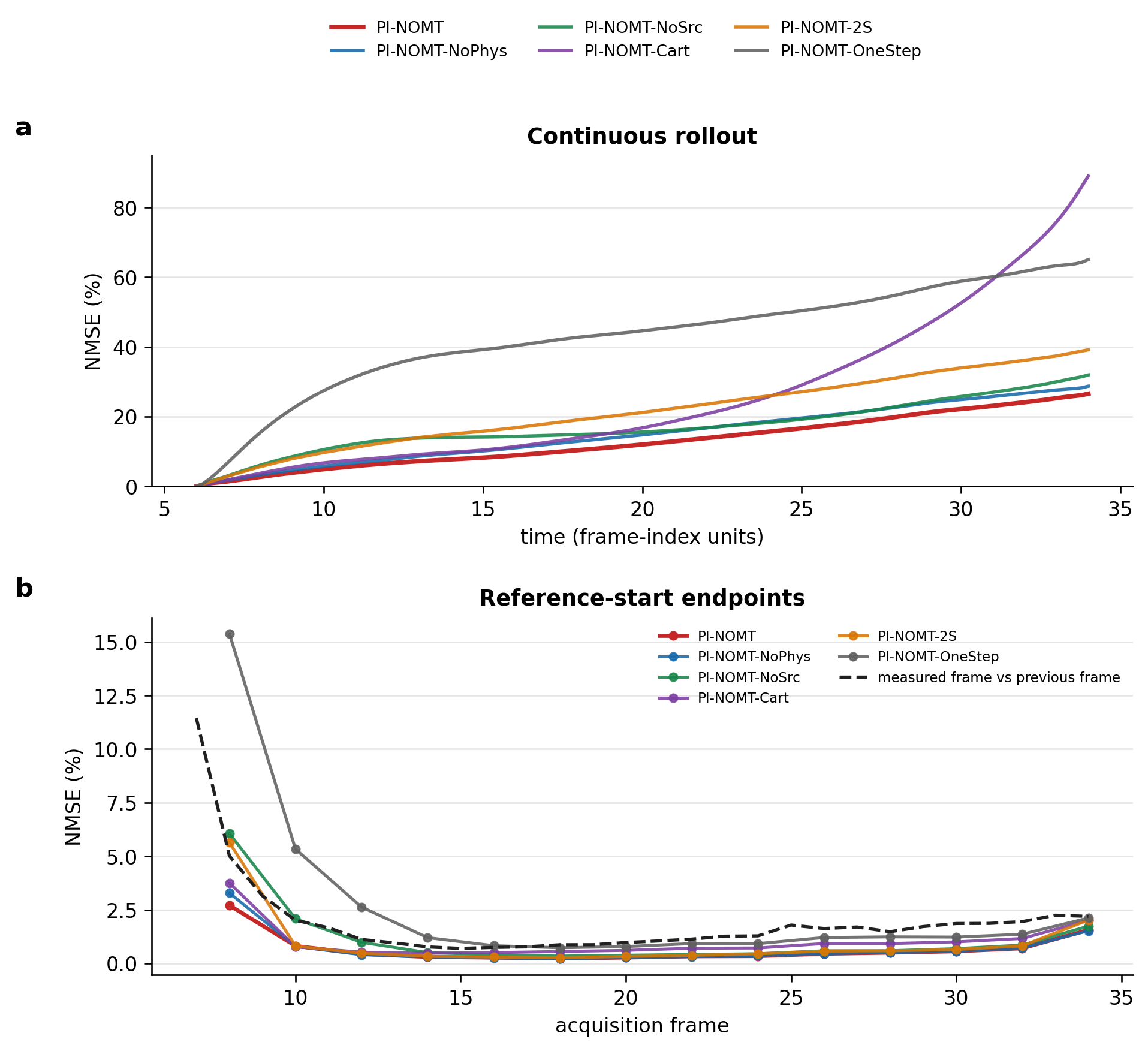}
\caption{Density-error curves for selected ablations.  Panel (a) measures
recursive rollout from frame 6.  The reference-start endpoint curve in panel
(b) measures two-frame chunks initialized from the teacher at the start of each
chunk and includes the intrinsic change between consecutive measured frames.
The large gap between
local and continuous error for PI-NOMT-Cart reveals recursive error
accumulation.}
\Description{Continuous-rollout and reference-start endpoint NMSE panels
compare six PI-NOMT variants.  PI-NOMT-Cart has
moderate local endpoint error but sharply increasing long-horizon error.}
\label{fig:selected_nmse_curves}
\end{figure*}

\begin{figure*}[!htbp]
\centering
\includegraphics[width=\textwidth]{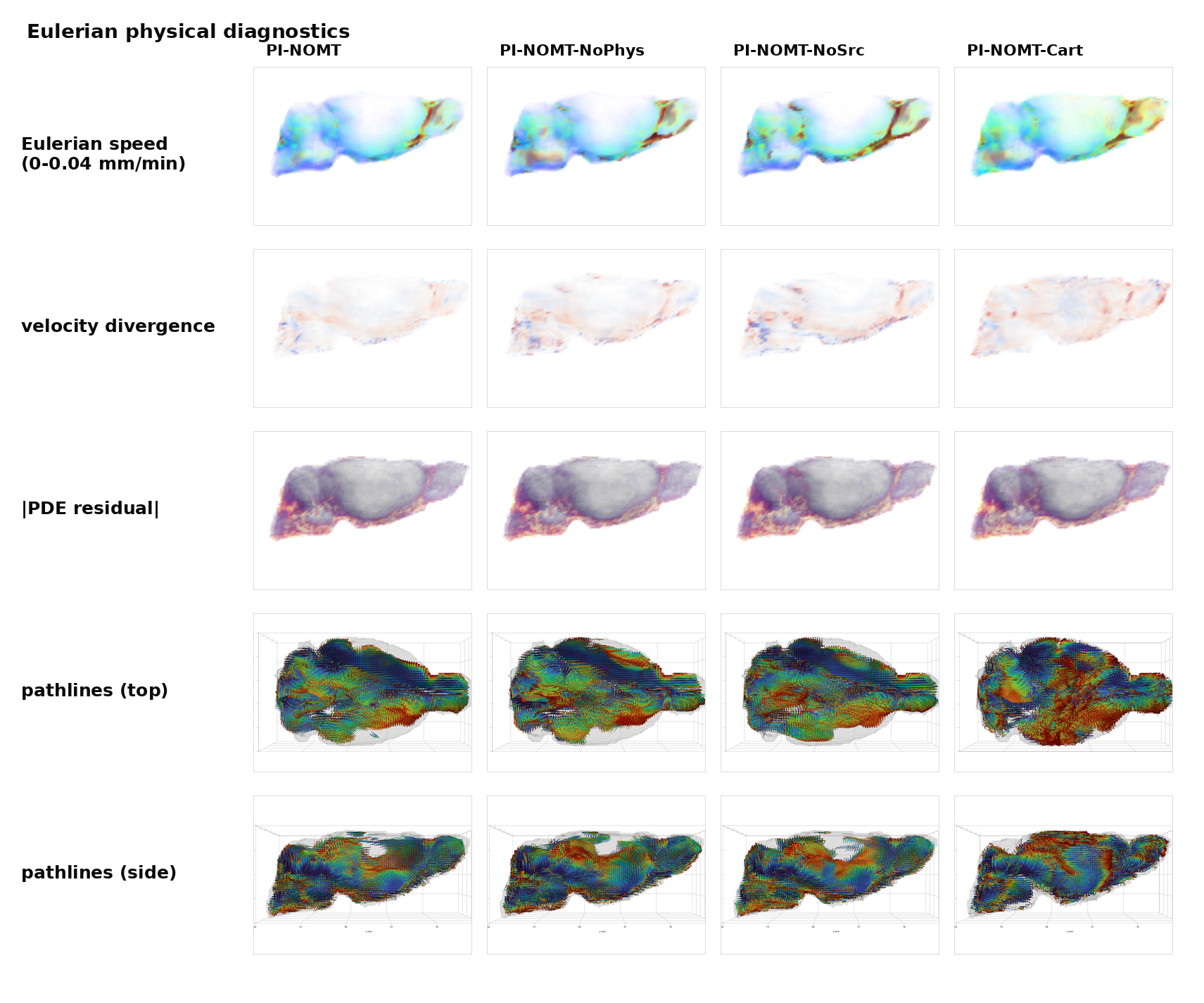}
\caption{Physical and Lagrangian diagnostics for PI-NOMT and three
ablations. Columns show PI-NOMT, PI-NOMT-NoPhys,
PI-NOMT-NoSrc, and PI-NOMT-Cart.
Rows show time-averaged speed, signed velocity divergence, mean absolute
incompressible-form PDE residual, top-view pathlines, and side-view pathlines.
Speed maps use the shared physical
range 0--0.04~mm/min, with larger values displayed at the upper color, and
the same combined opacity as the main-paper rat-brain transport summary,
$\alpha=\rho_n^{0.90}(0.01+0.99v_n)$.}
\Description{Five-by-four grid comparing Eulerian speed, velocity divergence,
PDE residual, and two pathline orientations for PI-NOMT and three ablations.}
\label{fig:physical_diagnostic_maps}
\end{figure*}

\begin{figure*}[!htbp]
\centering
\includegraphics[width=0.90\textwidth]{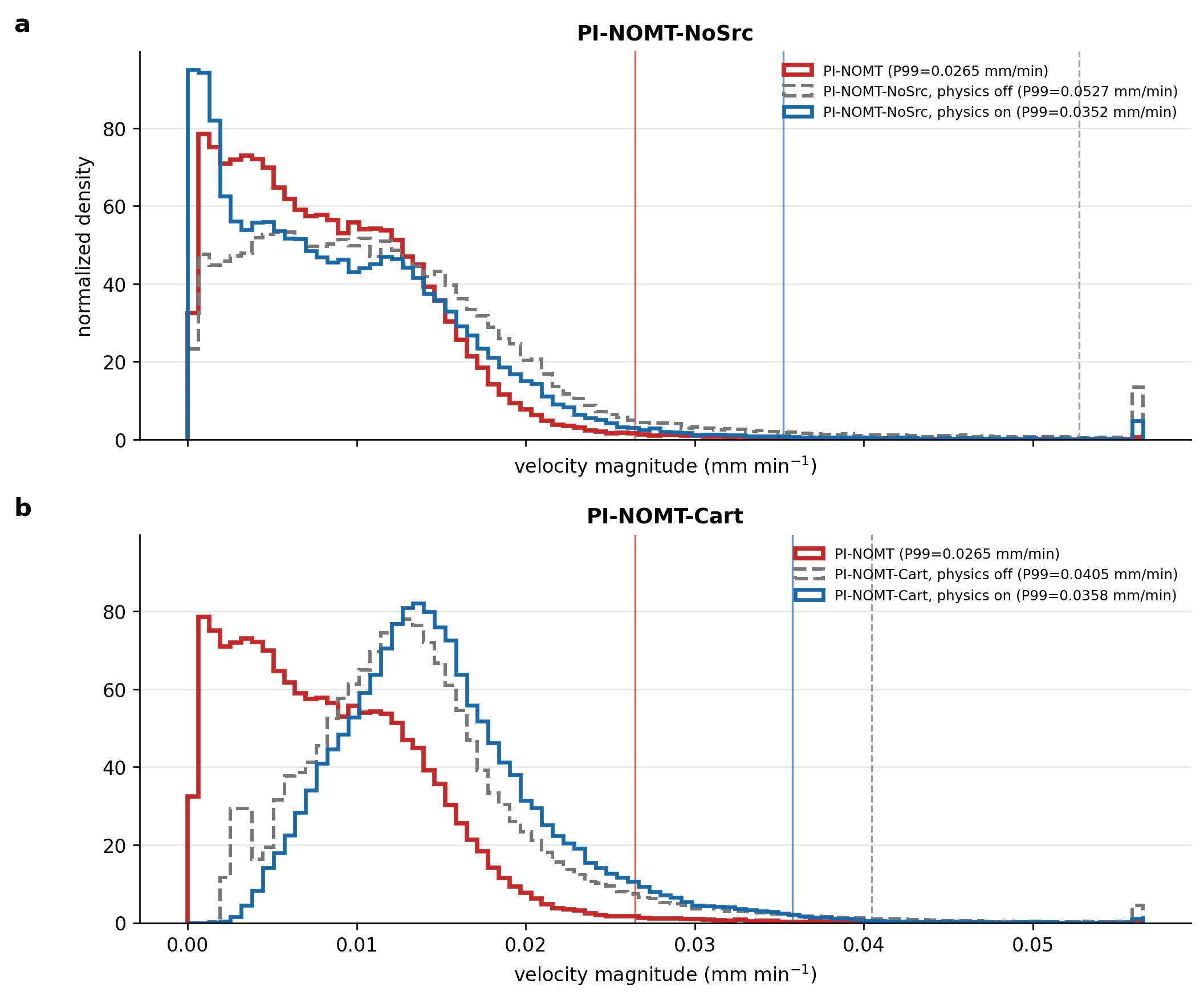}
\caption{Velocity-magnitude distributions for PI-NOMT and the two
structural ablations.  Panels compare physics-off and physics-on versions of
(a) PI-NOMT-NoSrc and (b) PI-NOMT-Cart against the selected PI-NOMT
reference.  Physics supervision contracts both high-speed tails, but their P99
values remain above that of PI-NOMT.}
\Description{Two overlaid velocity-magnitude histograms compare physics-off and
physics-on NoSrc and Cart models with PI-NOMT; legends report P99 values.}
\label{fig:velocity_histograms}
\end{figure*}

\FloatBarrier

\section{Cross-Subject Visualizations}
\label{supp:cross_subject_visualizations}

\begin{figure*}[!htbp]
\centering
\includegraphics[width=\textwidth]{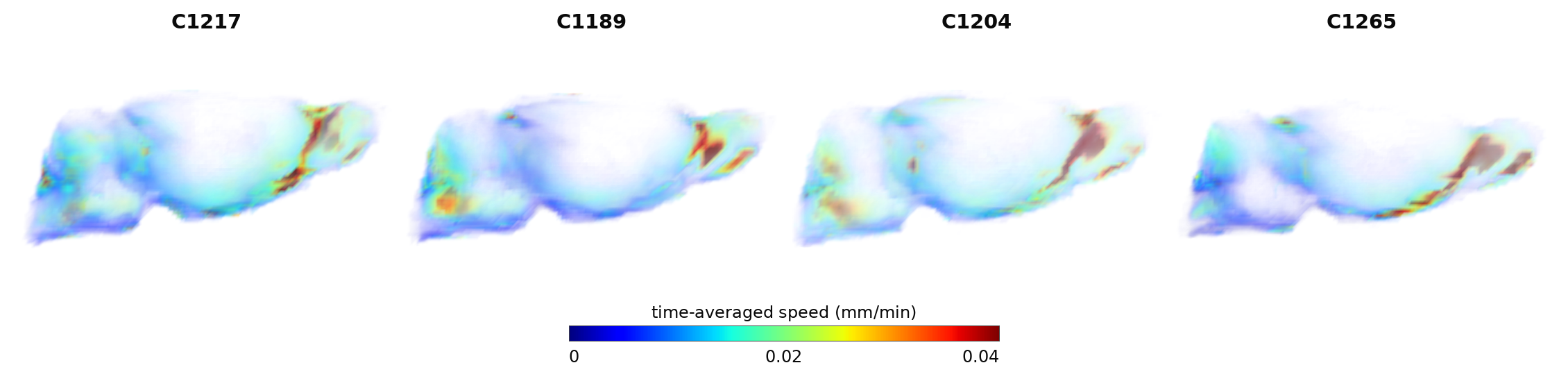}
\caption{Time-averaged Eulerian speed over frames 6--34 for the detailed C1217
experiment and three additional subjects.  All panels use the same
0--0.04~mm/min scale and the density--speed opacity mapping used in the main-paper rat-brain transport summary.}
\Description{Four side-view rat-brain maps compare time-averaged Eulerian speed
for subjects C1217, C1189, C1204, and C1265 under one shared color scale.}
\label{fig:cohort_speed}
\end{figure*}

\begin{figure*}[!htbp]
\centering
\includegraphics[width=\textwidth]{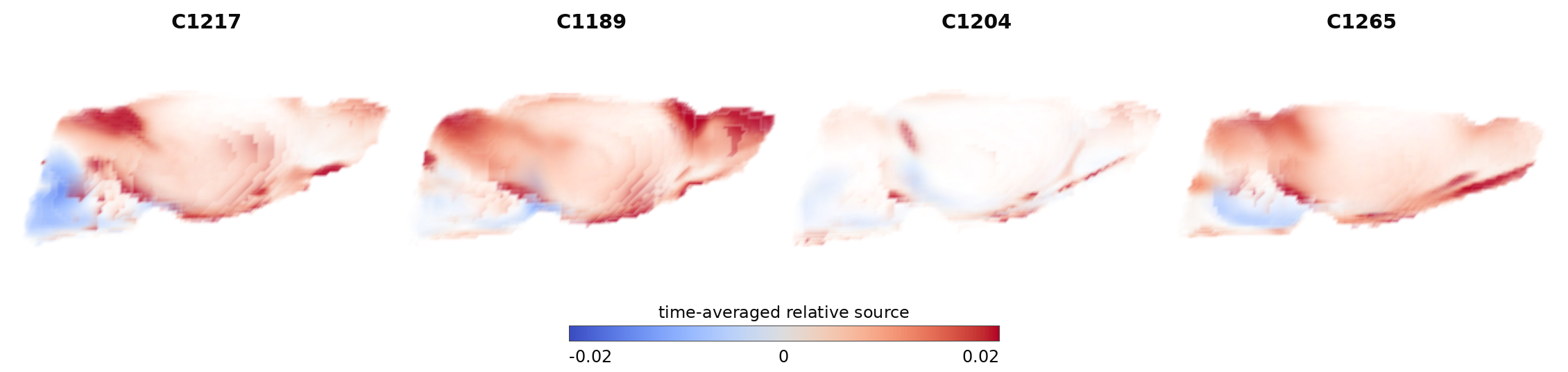}
\caption{Time-averaged Eulerian relative-source fields over frames 6--34 for
the same four subjects.  A common $[-0.02,0.02]$ scale and common
density--source opacity mapping are used throughout.}
\Description{Four side-view rat-brain maps compare signed time-averaged
relative-source fields for subjects C1217, C1189, C1204, and C1265.}
\label{fig:cohort_source}
\end{figure*}

\begin{figure*}[!htbp]
\centering
\includegraphics[width=\textwidth]{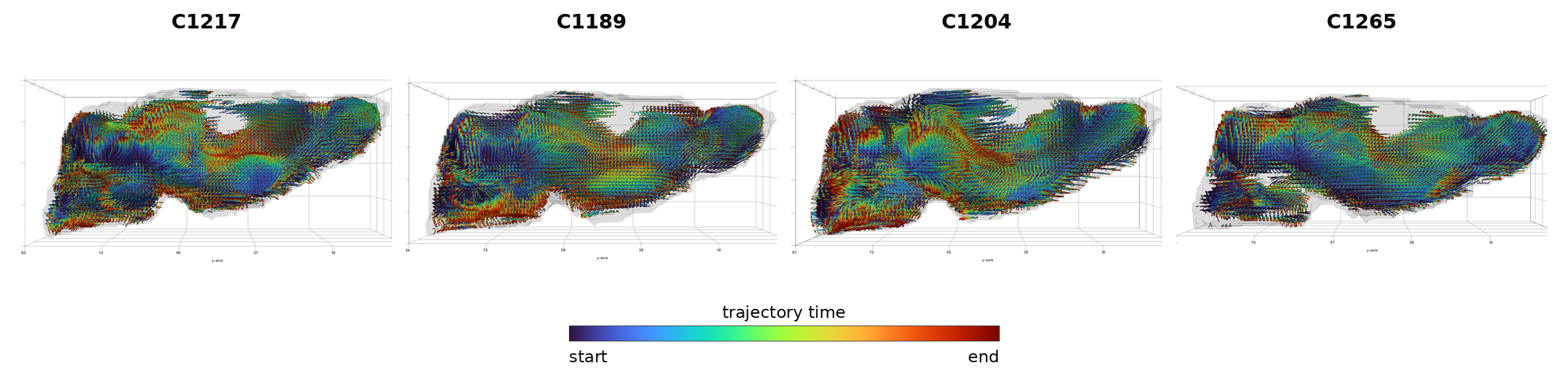}
\caption{Side-view pathlines generated using the diffusion-augmented drift for the four selected subjects.  Identical rendering settings are used, and color indicates progression from trajectory start to end.}
\Description{Four side-view pathline renderings show trajectory progression
for subjects C1217, C1189, C1204, and C1265.}
\label{fig:cohort_pathlines}
\end{figure*}

\begin{figure*}[!htbp]
\centering
\includegraphics[width=\textwidth]{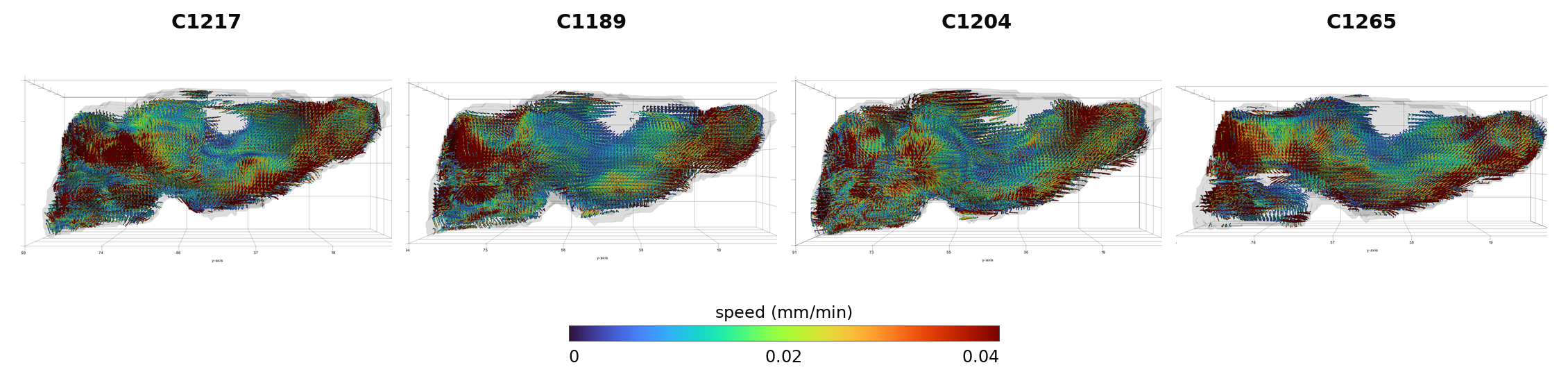}
\caption{Side-view speed-lines for the four selected subjects.  Trajectory points are colored by the local advective speed \(\|\mathbf{v}_{\boldsymbol{\theta}}\|_2\) using one shared 0--0.04~mm/min scale.}
\Description{Four side-view speed-line renderings compare instantaneous learned
speed for subjects C1217, C1189, C1204, and C1265 under one shared scale.}
\label{fig:cohort_speedlines}
\end{figure*}

\begin{figure*}[p]
\centering
\includegraphics[width=0.975\textwidth]{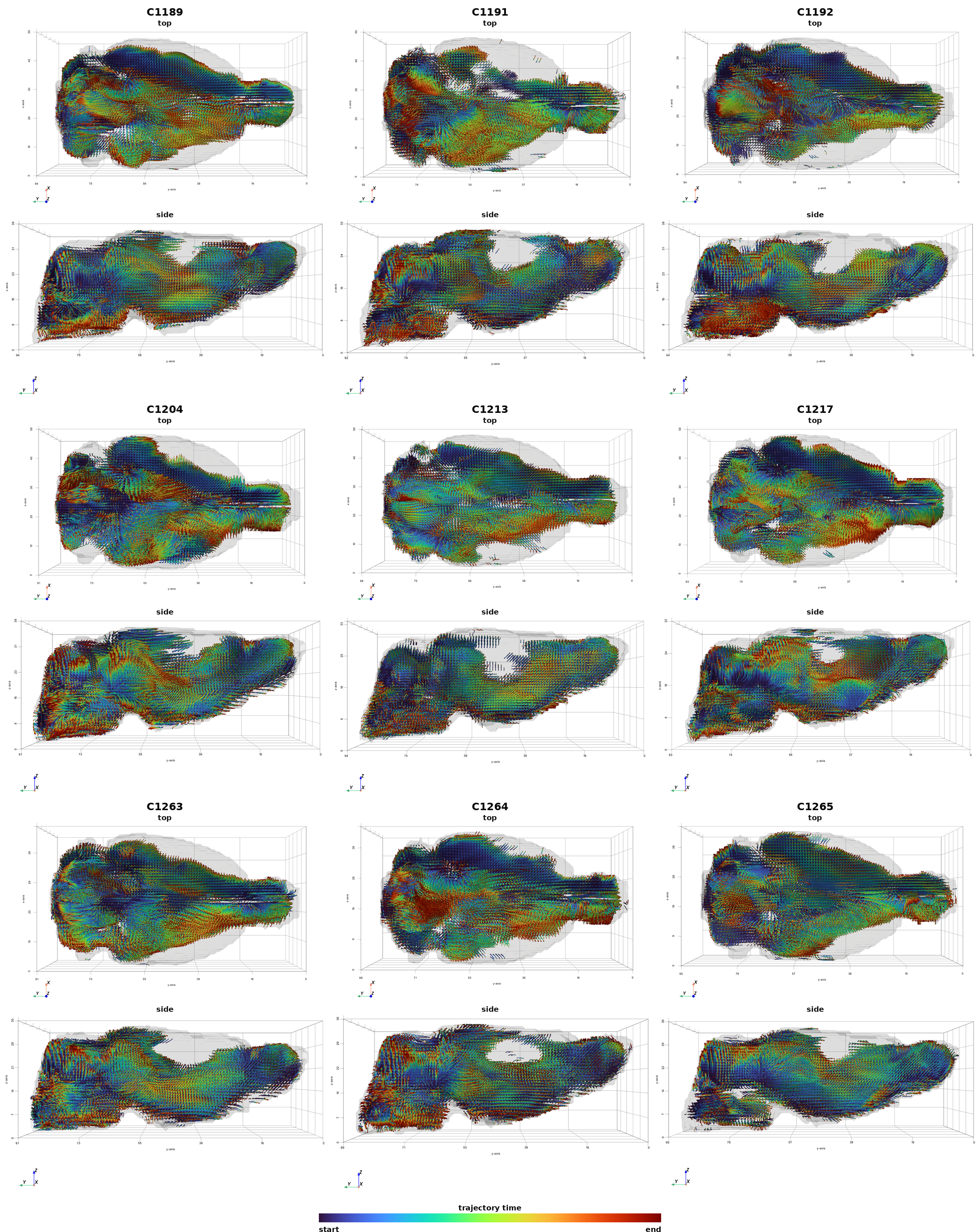}
\caption{Complete cross-subject pathline comparison. Top and side views are
shown for all nine subjects under common integration and rendering settings;
color indicates progression from trajectory start to end.}
\Description{A three-column arrangement shows stacked top and side pathline
views for all nine control-rat subjects under common rendering settings.}
\label{fig:cohort_appendix_pathlines}
\end{figure*}

\begin{figure*}[p]
\centering
\includegraphics[width=0.975\textwidth]{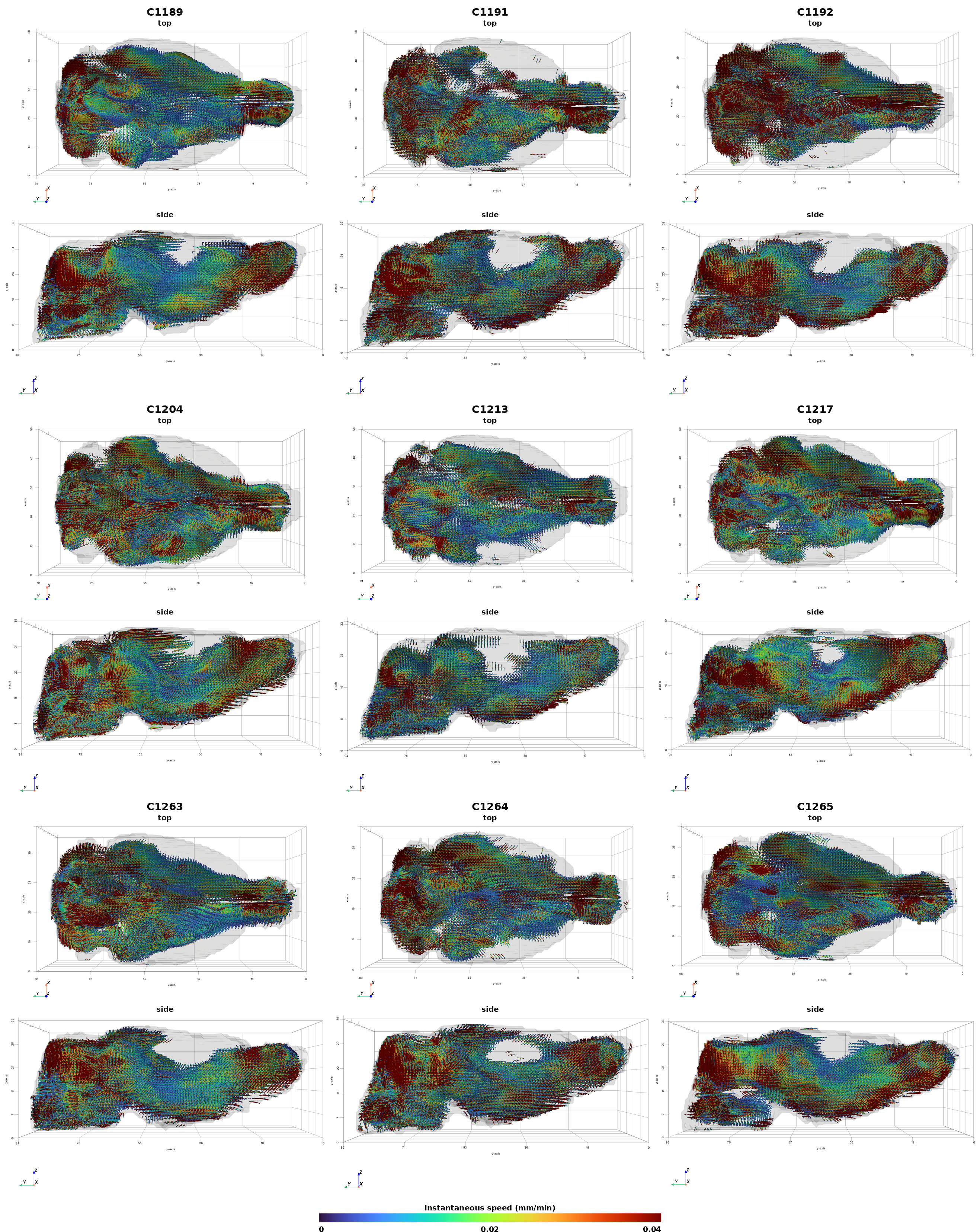}
\caption{Complete cross-subject speed-line comparison. Top and side views are
shown for all nine subjects; trajectory points are colored by instantaneous
learned speed using a shared 0--0.04~mm/min scale.}
\Description{A three-column arrangement shows stacked top and side speed-line
views for all nine control-rat subjects under a shared speed scale.}
\label{fig:cohort_appendix_speedlines}
\end{figure*}

\FloatBarrier


\end{document}